# Interpretable label-free self-guided subspace clustering

*Ivica Kopriva,*

*Division of Computing and Data Science*

*Ruđer Bošković Institute, Zagreb, Croatia*

**Abstract**

Majority subspace clustering (SC) algorithms depend on one or more hyperparameters that need to be carefully tuned for the SC algorithms to achieve high clustering performance. Hyperparameter optimization (HPO) is often performed using grid-search, assuming that some labeled data is available. In some domains, such as medicine, this assumption does not hold true in many cases. One avenue of research focuses on developing SC algorithms that are inherently free of hyperparameters. For hyperparameters-dependent SC algorithms, one approach to label-independent HPO tuning is based on internal clustering quality metrics (if available), whose performance should ideally match that of external (label-dependent) clustering quality metrics. In this paper, we propose a novel approach to label-independent HPO that uses clustering quality metrics, such as accuracy (ACC) or normalized mutual information (NMI), that are computed based on pseudo-labels obtained from the SC algorithm across a predefined grid of hyperparameters. Assuming that ACC (or NMI) is a smooth function of hyperparameter values it is possible to select subintervals of hyperparameters. These subintervals are then iteratively further split into halves or thirds until a relative error criterion is satisfied. In principle, the hyperparameters of any SC algorithm can be tuned using the proposed method. We demonstrate this approach on several single- and multi-view SC algorithms, comparing the achieved performance with their oracle versions across six datasets representing digits, faces and objects. The proposed method typically achieves clustering performance that is 5% to 7% lower than that of the oracle versions. We also make our proposed method interpretable by visualizing subspace bases, which are estimated from the computed clustering partitions. This aids in the initial selection of the hyperparameter search space.

## 1.0 Introduction

Clustering is a fundamental problem in unsupervised learning [1]. It has numerous applications, including medical image analysis [2, 3], single-cell transcriptomics [4], pattern recognition [5], speech recognition [6], image segmentation [7], and data mining [8], among others. Due to the complex shapes of samples spaces, distance-based clustering algorithms, such as k-means [9], often struggle to cluster data accurately in the original ambient domain. However, if a high-quality data-affinity matrix can be estimated, spectral methods can achieve high clustering performance [10-12]. Subspace clustering (SC) algorithms focus on learning a data affinity matrix under the assumption that the data are generated by a union-of-linear subspaces [13-17]. Since contemporary data are often recorded across multiple modalities or represented by various multiple features, multi-view extensions of SC algorithms have also been proposed [18-24]. However, real world data do not always originate from linear subspaces. To address this issue, SC algorithms can be formulated in a Reproducing Kernel Hilbert Space (RKHS), also known as the feature space, [25-27]. An alternative to kernel-based SC is the graph filtering approach [28-30]. As discussed in [28], graph filtering smooths the graph, removes noise, and iteratively incorporates graph similarity into features. This process can make data separable in the graph-filtering domain, even if they are not separable in the original space. As demonstrated in [29, 30], this approach can achieve performance comparable to that of deep methods. In this work, we combine graph filtering with a selected SC algorithm. Deep SC [31-36] is motivated by learning embeddings where the use of union-of-linear subspace model is more appropriate than in the original input space. Additionally, deep SC algorithms are valued for their powerful representation learning capabilities.

Although the SC algorithms cited, as well as many related ones, exhibit excellent clustering performance on benchmark datasets, all of them involve one or more hyperparameters. Hyperparameter optimization (HPO) in these algorithms is primarily based on external cluster quality metrics, such as clustering error, which require a certain amount of labeled data. While SC algorithms are designed to operate in a purely unsupervised manner, it is often assumed in practice that a validation subset with labeled samples is available [37]. However, real-world clustering tasks frequently lack label information to aid in hyperparameter selection [38]. For instance, in the medical field, the number of labeled data is limited, and human annotation is both time consuming and expensive, even though a large number of unlabeled data is available [39, 40]. Consequently, there is a growing interest in self-supervised learning (SSL) algorithms [39, 40], which aim to learn useful features from a large number of unlabeled instances without relying on human annotation.

One approach to label-free learning involves a predefined pretext task, which is a key concept in SSL [39-41]. A pretext task is designed for a deep learning algorithm to solve, generating pseudo labels in the process. A commonly used pretext task in context-based methods is rotation [42, 39, 40]. In this approach, each original image is rotated by $90^0$, $180^0$, and $270^0$, resulting in four classes: $0^0$, $90^0$, $180^0$, and $270^0$. The performance of the model on this self-supervised pretext task is crucially linked to its performance on downstream supervised recognition tasks. Another label-free tuning approach for SC algorithms involves using internal clustering quality metrics [43]. It has been noted that existing metrics for clustering quality [44-47] are not suitable for evaluating the internal clustering quality of union-of-linear-subspaces models. Extensive validation in [43] has demonstrated that the *K*-subspaces (KSS) cost used in the KSS algorithm [48-50] and the Calinski-Harabasz (CS) index [51] are effective for hyperparameters selection when the number of clusters is known in advance. These metrics require three inputs: the data, the estimated clusters and set of subspace dimensions. The last input might be challenging to provide when working with datasets in new application domains.

It has been noted [52] that the performance of machine learning (ML) algorithms is largely determined by the hyperparameter settings used. The main challenge is that hyperparameters must be tuned for each specific ML problem to achieve optimal performance. Consequently, HPO, which aims to find the best configuration for ML tasks, is a significant area of research topics in the ML community. The most commonly used strategy is search-based, where a predefined search space is used to find the optimal hyperparameter values for a given ML algorithm [53-57]. These approaches are computationally intensive and require labeled data to evaluate performance for different hyperparameter values. An efficient alternative is meta-learning (MtL) [58], which uses previous evaluations from historical datasets to predict desirable hyperparameters for new task. Traditionally, MtL has worked well for hyperparameters organized as vectors. However, [52] proposes organizing multiple hyperparameters as tensors and formulates the interpolation of optimal hyperparameter values as a low-rank tensor completion (LRTC) problem [59-61]. This approach assumes that the selected performance metric is a smooth function of the hyperparameters, allowing for interpolation of optimal values from historic evaluations through solving the LRTC problem. A critical assumption of this approach is the availability of previous historic evaluations. In rapidly evolving domains, such as various medical imaging modalities, where new data are generated frequently, it is often unrealistic to rely on such historical evaluations.

To address the hyperparameter tuning challenges outlined above, we propose a new HPO strategy for derived SC algorithms. Unlike approaches in [39-41], which depend on the definition or selection of a pretext task, or [43], which requires choosing an appropriate internal clustering quality metric (if possible), we introduce a label-free HPO method for SC algorithms. This method is based on clustering quality metrics such as accuracy (ACC) or normalized mutual information (NMI). However, instead of using external (hard) labels, ACC and NMI in our approach are calculated from pseudo-labels generated by the SC algorithm itself. Similar to the grid-search approach for HPO, we define a search space for selected SC algorithm where the optimal hyperparameters are expected to reside. However, this space can be less dense compared to an exhaustive greed search. In our approach, as in [52], we also assume the performance metric is a smooth function of the hyperparameters. Accordingly, we compute ACC or NMI between pseudo-labels generated by neighboring hyperparameter values. Based on the smoothness assumption, we subdivide hyperparameter intervals into smaller sections, which are further split into halves or thirds, and SC algorithm generates pseudo-labels for these interpolated values. This process is repeated iteratively until a relative error criterion is met. Thus, our HPO approach allows SC algorithms to be tuned in a label-free manner, enabling their application in new domains where labeled data for HPO is unavailable. Furthermore, our approach complements an existing avenue of research related to development of SC algorithms that are free of hyperparameters [20-23, 37, 51, 63, 64]. In other words, our approach is proposed for label-free self-guided (LFSG) hyperparameter tuning of existing SC algorithms. Figure 1 illustrates our approach to HPO using as an example the least squares regression (LSR) SC algorithm [17]. MATLAB code of the proposed approach to Label-Free Self-Guided Subspace Clustering is available at https://github.com/ikopriva/LFSGSC.

In fields like medicine, achieving high diagnostic performance often requires the use of highly complex models whose decision-making processes are challenging to interpret and explain [65, 66]. When decisions involve high-stakes, as medical diagnosis, it becomes crucial to provide explanations for an algorithm's predictions. To the best of our knowledge, SC algorithms also face this issue of interpretability. In response, we propose a method to interpret clustering results from SC algorithms by visualizing subspace bases estimated from the obtained clustering partitions. If experts judge the visualization quality to be inadequate, the initial hyperparameter search space can be refined, and the algorithm restarted.

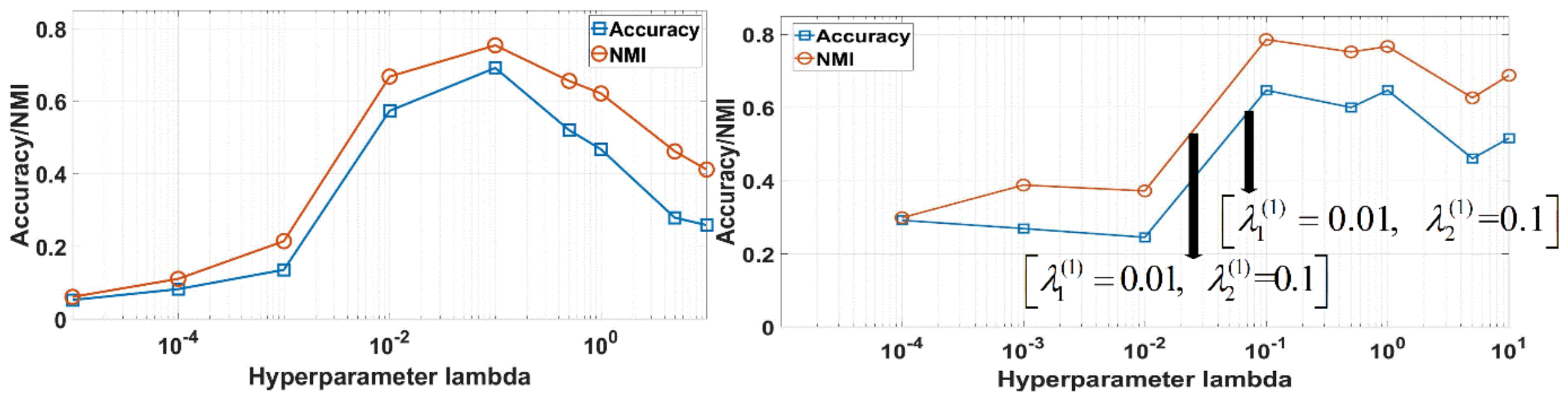

(**a**) max ACC($\lambda_5$=0.1)=0.692
max NMI($\lambda_5$=0.1)=0.754

(**b**) max ACC($\lambda_4 = 0.01, \lambda_5$=0.1)=0.642
max NMI($\lambda_4 = 0.01, \lambda_5$=0.1)=0.808

**Figure 1**. Illustration of proposed approach to label-free HPO for LSR SC algorithm [17], that depends on one hyperparameter λ, see eq.(6), on Extended YaleB dataset. (**a**) Accuracy and NMI at $\lambda \in \{10^{-5}, 10^{-4}, 10^{-3}, 10^{-2}, 10^{-1}, 0.5, 1, 5, 10\}$ calculated with external (hard) labels. Both maximal accuracy and maximal NMI occur at $\lambda_5$=0.1. (**b**) Accuracy and NMI computed between pseudo-labels generated by the same grid values of λ as in (**a**). Maximal accuracy occurs between $\lambda_4$=0.01 and $\lambda_5$=1. Consequently, new iteration "1" begins on the interval with borders $\lambda_1^{(1)} = \lambda_4 = 0.01$ and $\lambda_2^{(1)} = \lambda_5 = 0.1$. The interval is further subdivided into thirds giving $\{0.01, 0.04, 0.07, 0.1\}$ and the process is repeated until convergence is reached. Regarding NMI, the process is analogous to the accuracy metric.

Our contributions in this paper are summarized as follows:

- We propose a label-free approach to HPO for existing SC algorithms. This approach does not rely on defining a pretext task that must be correlated with downstream SC clustering task, nor does it depend on internal clustering quality metrics, which can be difficult to determine for certain SC algorithms or datasets. Additionally, it does not require prehistoric evaluations, which may not be available for data from some domains. Instead, our approach uses clustering quality metrics such as ACC or NMI. However, rather than relying on actual labels, ACC or NMI are computed from pseudo-labels generated by the selected SC algorithm within a predefined search space, where the optimal hyperparameter values are expected to reside. Our approach assumes that ACC and NMI are smooth functions of hyperparameters, allowing for the selection of the subintervals where ACC and NMI values are maximized. These intervals are further split into halves or thirds, and the process is repeated until the relative error between ACC and NMI values and the hyperparameter values is less than or equal to a predefined threshold. Our approach allows for the re-use of existing SC algorithms in domains where labeled data is unavailable.

- We propose a method to interpret clustering results from SC algorithms through the visualization of subspace bases, estimated from the obtained clustering partitions. If the visualization quality is deemed insufficient, the initial hyperparameter search space can be adjusted, and the algorithm restarted.

- We evaluate the proposed approach to label-free HPO on several carefully selected SC algorithms and compare the resulting performance with the oracle versions of the same algorithms. Specifically, we demonstrate the effectiveness of our approach on linear single-view and multi-view SC algorithms, as well as on some nonlinear, kernel-based and graph-filtering-based versions. As shown, our approach achieves performance close to that of the oracle versions of the same SC algorithms.

- As side contributions, we formulate graph-filtering version of the LSR SC algorithm in Algorithm 1, and out-of-sample extension of kernel-based version of the LSR SC algorithm in section 3.4.

## 2.0 Background and Related Work

### *2.1. Subspace clustering*

We assume that the columns of the matrix $\mathbf{X} \in \mathbb{R}^{D\times N}$ represent data points drawn from a union of linear subspaces $\bigcup_{i=1}^{C} S_i$ of unknown dimensions $\left\{d_i = \dim\left(S_i\right)\right\}_{i=1}^{C}$ in $\mathbb{R}^D$ where:

$$S_i = \left\{\mathbf{x}_n \in \mathbb{R}^{D\times 1} : \mathbf{x}_n = \mathbf{A}_i \mathbf{z}_n^i \right\}_{n=1}^{N} \quad i \in \{1,...,C\} \tag{1}$$

where $C$ represents number of subspaces, which coincides with the number of clusters, $\left\{\mathbf{A}_i \in \mathbb{R}^{D\times d_i}\right\}_{i=1}^{C}$ represents bases of subspaces, $\left\{d_i \ll D\right\}_{i=1}^{C}$ represents dimensions of subspaces, and $\left\{\mathbf{z}_n^i \in \mathbb{R}^{d_i}\right\}_{i=1}^{C}$ stands for representations of data samples $\left\{\mathbf{x}_n\right\}_{n=1}^{N}$. In the most general formulation of SC problem, the goal is to identify the number of subspaces, the bases of these subspaces, their dimensions, and to cluster the data points according to the subspaces from which they are generated [13]. It is typically assumed that the number of subspaces $C$ is known *a priori*. Moreover, SC primarily focuses on clustering [14-17]. Under this specific data model, SC algorithms aim to learn the representation matrix $\mathbf{Z} \in \mathbb{R}^{N\times N}$, from which the data affinity matrix $\mathbf{W} \in \mathbb{R}^{N\times N}$ is derived as follows:

$$\mathbf{W} = \frac{|\mathbf{Z}| + |\mathbf{Z}|^{\mathrm{T}}}{2} \tag{2}$$

The diagonal degree matrix **D**, based on the affinity matrix **W**, is computed as:

$$d_{ii} = \sum_{j=1}^{N} w_{ij} \quad i \in \{1,..,N\} \ . \tag{3}$$

Using equations (2) and (3), the normalized graph Laplacian matrix is computed as [11]:

$$\mathbf{L} = \mathbf{I} - (\mathbf{D})^{-1/2}\, \mathbf{W}(\mathbf{D})^{-1/2} \tag{4}$$

The spectral clustering algorithm [10, 11], is then applied to **L** to assign cluster labels to the data points: $\mathbf{F} \in \mathbb{N}_0^{N\times C}$. Based on the self-expressive data model **X**=**XZ**, many SC algorithms focus on learning the representation matrix **Z** by solving the following optimization problem:

$$\min_{\mathbf{Z}} \tfrac{1}{2}\left\|\mathbf{X} - \mathbf{XZ}\right\|_F^2 + \lambda f(\mathbf{Z}) + \tau g(\mathbf{Z}) \quad \text{s.t.} \quad \mathrm{diag}(\mathbf{Z}) = \mathbf{0}. \tag{5}$$

In equation (5), $f$ and $g$ are regularization functions imposed on **Z**, while $\lambda$ and $\tau$ are regularization constants (hypeparameters). For low-rank representation SC (LRR SC) [14], $f(\mathbf{Z}) = \|\mathbf{Z}\|_*$ and $\tau$=0. For sparse SC (SSC) [15], $f(\mathbf{Z}) = \|\mathbf{Z}\|_1$ and $\tau$=0. For LSR SC [17], $f(\mathbf{Z}) = \|\mathbf{Z}\|_F$ and $\tau$=0. For S0L0 low-rank sparse SC (LRSSC) [16], $f(\mathbf{Z}) = \|\mathbf{Z}\|_{S_0}$ and $g(\mathbf{Z}) = \|\mathbf{Z}\|_0$. If the constraint diag(**Z**)=**0** is removed from equation (5), the LSR SC algorithm has an analytical solution:

$$\mathbf{Z} = \left(\mathbf{X}^{\mathrm{T}}\mathbf{X} + \lambda\mathbf{I}\right)^{-1}\mathbf{X}^{\mathrm{T}}\mathbf{X}. \tag{6}$$

For LRR SC, SSC, and S0L0 LRSSC solutions of equation (5) are obtained iteratively using the alternating direction method of multipliers (ADMM) [67].

It is straightforward to derive the nonlinear kernel-based version of the LSR SC algorithm (6) by replacing $\mathbf{X}^{\mathrm{T}}\mathbf{X}$ with the Gramm matrix $\mathbf{K}(\mathbf{X},\mathbf{X})$:

$$\mathbf{Z} = \left(\mathbf{K}(\mathbf{X},\mathbf{X}) + \lambda\mathbf{I}\right)^{-1}\mathbf{K}(\mathbf{X},\mathbf{X}). \tag{7}$$

Thus, the kernel version of the LSR SC also depends on one hyperparameter, $\lambda$. However, the number of hyperparameters increases depending on the choice of the kernel function. For instance, if we select the Gaussian kernel $\kappa(\mathbf{x}_i,\mathbf{x}_j) = \exp\left(-\|\mathbf{x}_i - \mathbf{x}_j\| / 2\sigma^2\right)$, the variance $\sigma^2$ becomes an additional hyperparameter.

Graph filtering can be an alternative to kernel methods for clustering data generated from manifolds [68]. This method generates a smoothed (filtered) version of the feature matrix as follows:

$$\bar{\mathbf{X}}^{\mathrm{T}} = \left(\mathbf{I} - \frac{\mathbf{L}}{2}\right)^{k}\mathbf{X}^{\mathrm{T}} \tag{8}$$

where $k$ is a non-negative integer. Instead of applying the selected SC algorithm to the original data matrix **X**, it is applied to the graph-filtered data $\bar{\mathbf{X}}$. The data adjacency matrix **W** is typically computed from equation (2), where **Z** is the representation matrix estimated by a SC using self-representation model **X=XZ**. Since the goal is to apply SC algorithm to $\bar{\mathbf{X}}$, we implement a graph-filtering version of the algorithm in an iterative manner [68]. This process is illustrated in Algorithm 1, which combines graph filtering with the LSR SC. Unlike kernel methods, it does not suffer from problems analogous to kernel selection.

**Algorithm 1** Graph filtering least squares subspace clustering

**Inputs**: Feature (data) matrix **X**, stopping criterion $\varepsilon$=10^-4^.

**Parameters**: Filter order $k$, $\lambda$-regularization constant

1: Initialize $t$=0 and $\bar{\mathbf{X}}_1 = \mathbf{X}$

2: **repeat**

3: $t \leftarrow t+1$

4: $\mathbf{Z}_t \leftarrow \left(\bar{\mathbf{X}}_t^{\mathrm{T}}\bar{\mathbf{X}}_T + \lambda\mathbf{I}\right)^{-1}\bar{\mathbf{X}}_t^{\mathrm{T}}\bar{\mathbf{X}}_t$

5: $\mathbf{W}_t = \frac{1}{2}\left(|\mathbf{Z}_t| + |\mathbf{Z}_t|^{\mathrm{T}}\right)$

6: $\mathbf{L}_t = \mathbf{I} - \mathbf{D}_t^{-1/2}\mathbf{W}_t\mathbf{D}_t^{-1/2}$

7: $\bar{\mathbf{X}}_{t+1}^{\mathrm{T}} = \left(\mathbf{I} - \frac{\mathbf{L}_t}{2}\right)^k \mathbf{X}^{\mathrm{T}}$

8: **until** $\|\mathbf{W}_t - \mathbf{W}_{t-1}\|_F^2 \le \varepsilon$

**Output**: Graph-filtered adjacency matrix $\mathbf{W} \leftarrow \mathbf{W}_t$

In many real-world experiments, collected data originate from multiple modalities. For example, in PET-CT imaging, co-registered PET and CT images are recorded simultaneously [69]. Similarly, the same documents may be available in multiple languages [70], while different descriptors can be generated for the same image [71]. These scenarios lead to the multi-view SC (MVSC) problem [19-24, 71, 72]. A multi-view dataset composed of $V$ views is denoted as $\left\{\mathbf{X}^{(v)} \in \mathbb{R}^{D_v \times N}\right\}_{v=1}^{N}$. The objective of MVSC algorithms is to learn a data affinity matrix that integrates all the views and then apply spectral clustering to assign labels to data points. The MVSC method proposed in [24], known as LMVSC, has linear complexity in terms of the number of samples and solves the following optimization problem:

$$\min_{\mathbf{Z}^{(v)}} \sum_{v=1}^{V} \left\| \mathbf{X}^{(v)} - \mathbf{A}^{(v)} \left(\mathbf{Z}^{(v)}\right)^{\mathrm{T}} \right\|_F^2 + \lambda \left\|\mathbf{Z}^{(v)}\right\|_F^2 \ \text{ s.t.} \mathbf{0} \le \mathbf{Z}^{(v)}, \left(\mathbf{Z}^{(v)}\right)^{\mathrm{T}}\mathbf{1} = \mathbf{1}. \tag{9}$$

In equation (9), $\mathbf{A}^{(v)} \in \mathbb{R}^{D_v \times M}$ represents the $M$ anchors for the $v$-th view, $\mathbf{Z}^{(v)} \in \mathbb{R}^{N \times M}$ is the representation matrix for the $v$-th view, and **1** is a vector of ones. The view-dependent data adjacency matrices are constructed as:

$$\mathbf{W}^{(v)} = \hat{\mathbf{Z}}^{(v)} \left( \hat{\mathbf{Z}}^{(v)} \right)^{\mathrm{T}} \quad \hat{\mathbf{Z}}^{(v)} = \mathbf{Z}^{(v)} \mathbf{\Sigma}^{-1/2}, \ \mathbf{\Sigma}_{ii} = \sum\nolimits_{j=1}^{N} \mathbf{Z}_{ji}^{(v)} \ . \tag{10}$$

The final data adjacency matrix that represents all views is obtained as:

$$\mathbf{W} = \frac{1}{V} \sum_{v=1}^{V} \mathbf{W}^{(v)} \ . \tag{11}$$

The outlined LMVCS algorithm has two hyperpamerters: $M$ and $\lambda$.

Almost all (traditional) SC algorithms follow a two-stage process to assign cluster labels to data points. In the first stage, spectral decomposition of data affinity matrix is performed, resulting in a continuous relaxation of the binary cluster indicator matrix. In the second stage, the *k*-means algorithm is used to round this continuous relaxation into a binary indicator matrix. However, this two-stage approach can lead to severe information loss and performance degradation [73, 72]. To address this issue, the multi-view clustering via multiple Laplacian embeddings (MCMLE) method was proposed in [72]. It aims to learn a unified $N \times C$ binary indicator clustering matrix $\mathbf{Y} \in Ind$ that represents all views. This matrix is obtained by solving the following optimization problem:

$$\begin{aligned} &\min_{\mathbf{F}^{(1)},\ldots,\mathbf{F}^{(V)},\mathbf{Y},\boldsymbol{\lambda}} \sum_{v=1}^{V} \lambda^{(v)} \left( tr\left( \mathbf{F}^{(v)\mathrm{T}} \mathbf{L}^{(v)} \mathbf{F}^{(v)} \right) - \alpha tr\left( \mathbf{F}^{(v)\mathrm{T}} \mathbf{D}^{(v)\frac{1}{2}} \mathbf{Y} \right) \right) + \beta \left\| \boldsymbol{\lambda} \right\|_2^2, \\ &\text{s.t.} \mathbf{F}^{(v)\mathrm{T}} \mathbf{F}^{(v)} = \mathbf{I}, \mathbf{Y} \in Ind, \sum_{v=1}^{V} \lambda^{(v)} = 1, \lambda^{(v)} \geq 0 \end{aligned} \quad , \tag{12}$$

where $\alpha$>0 and $\beta$>0 are penalty parameters, $\left\{ \mathbf{L}^{(v)} \right\}_{v=1}^{V}$ are view-specific normalized Laplacians in the form of equation (4), $\left\{ \mathbf{F}^{(v)} \in \mathbb{R}^{N \times C} \right\}_{v=1}^{V}$ are view-specific partitions, and components of $\boldsymbol{\lambda} \in \mathbb{R}_{0+}^{V \times 1}$ are view-specific weights. The MCLME algorithm introduces two hyperparameters: $\alpha$ and $\beta$.

**3.0 The proposed method**

In this section, we present our label-free approach to HPO for the selected SC algorithm. For the sake of readability, we first describe methodology for SC algorithms that depend a single hyperparameter. Examples of such algorithms include LSR SC (6), sparse SC [15], low-rank representation SC [14], LMVSC (9)-(11) [24]. We then extend our approach to SC algorithms that depend on two hyperparameters. Examples in this category include kernel LSR SC (7), assuming a Gaussian kernel, S0L0 LRSSC algorithm [16], graph filtering LSR SC [68] also presented in Algorithm 1, and the MCLME algorithm [72].

### *3.1. Label-free self-guided hyperparameter optimization*

Let us assume that the selected SC algorithm depends on a single hyperparameter $\lambda$, and that the predefined search space $\boldsymbol{\lambda} := [\lambda_1, ..., \lambda_M]$, $\boldsymbol{\lambda} \geq \mathbf{0}$, contains the optimal value $\lambda^*$. We also assume the hyperparameter values are ordered such that $\lambda_1 < \lambda_2 < ... < \lambda_M$. Next, we define the performance metrics to be used. First, we define accuracy (ACC) as the percentage of samples correctly labeled. Let the cluster labels (pseudo-labels) generated by the SC algorithm for hyperparameter values $\boldsymbol{\lambda}$ be denoted by:

$$\left\{ \mathbf{y}(\lambda_i) \in Ind := \left\{ y_n(\lambda_i) \right\}_{n=1}^{N} \right\}_{i=1}^{M} \quad . \tag{13}$$

ACC is then calculated as:

$$ACC\left(\mathbf{y}(\lambda_i), \mathbf{y}(\lambda_{i+1})\right) = \frac{\sum_{n=1}^{N} f\left(y_n(\lambda_i), y_n(\lambda_i + 1)\right)}{N} \times 100\% \;\; 1 \leq i < M \quad . \tag{14}$$

$f(a,b)$ is implemented as follows:

$$f(a,b) = \begin{cases} 1, & \text{if } a = \text{b} \\ 0, & \text{otherwise}. \end{cases}$$

Next, we define normalized mutual information (NMI), which measures the similarity between the estimated cluster labels for hypeparameter $\boldsymbol{\lambda}$. Using the same notation as for ACC, the MI between $\mathbf{y}(\lambda_i)$ and $\mathbf{y}(\lambda_{i+1})$ is described as:

$$MI\left(\mathbf{y}(\lambda_i), \mathbf{y}(\lambda_{i+1})\right) = \sum_{y_n \in \mathbf{y}(\lambda_i), y_m \in \mathbf{y}(\lambda_{i+1})} p(y_n, y_m) \log \frac{p(y_n, y_m)}{p(y_n)\, p(y_m)}$$

where $p(y_n)$ and $p(y_m)$ represent probabilities that any sample belongs to $\mathbf{y}(\lambda_i)$ and $\mathbf{y}(\lambda_{i+1})$, respectively. $p(y_n, y_m)$ is the probability that samples belong to both cluster sets $\mathbf{y}(\lambda_i)$ and $\mathbf{y}(\lambda_{i+1})$. NMI is defined as follows:

$$NMI(\mathbf{y}(\lambda_i), \mathbf{y}(\lambda_{i+1})) = \frac{MI(\mathbf{y}(\lambda_i), \mathbf{y}(\lambda_{i+1}))}{\sqrt{g(\mathbf{y}(\lambda_i))\, g(\mathbf{y}(\lambda_{i+1}))}} \times 100\% \; 1 \le i < M \; . \tag{15}$$

where $g(\mathbf{y}(\lambda_i))$ and $g(\mathbf{y}(\lambda_{i+1}))$ stand for entropies of $\mathbf{y}(\lambda_i)$ and $\mathbf{y}(\lambda_{i+1})$. They are defined as follows:

$$g(\mathbf{y}(\lambda_i)) = \sum_{n=1}^{N} \frac{y_n(\lambda_i)}{N} \log \frac{y_n(\lambda_i)}{N}$$

$$g(\mathbf{y}(\lambda_{i+1})) = \sum_{n=1}^{N} \frac{y_n(\lambda_{i+1})}{N} \log \frac{y_n(\lambda_{i+1})}{N} \; .$$

To simplify notation going forward, we shall use the shorthand $\mathbf{y}_i = \mathbf{y}(\lambda_i)$ and $\mathbf{y}_{i+1} = \mathbf{y}(\lambda_{i+1})$. Let $h(\mathbf{y}_i, \mathbf{y}_{i+1})$ represent either $ACC(\mathbf{y}_i, \mathbf{y}_{i+1})$ or $NMI(\mathbf{y}_i, \mathbf{y}_{i+1})$. By treating $\lambda_{i+1}$ as a constant, we assume that *h* is either a monotonically increasing function of $\lambda_i$ and $\lambda_{i+2}$:

$$h(\mathbf{y}_i, \mathbf{y}_{i+1}) \le h(\mathbf{y}_{i+1}, \mathbf{y}_{i+2}) \tag{16}$$

or a monotonically decreasing function of $\lambda_i$ and $\lambda_{i+2}$:

$$h(\mathbf{y}_i, \mathbf{y}_{i+1}) \ge h(\mathbf{y}_{i+1}, \mathbf{y}_{i+2}) \; . \tag{17}$$

Next, we now locate the subinterval $[\lambda_i, \lambda_{i+1}]$ where $h(\mathbf{y}_i, \mathbf{y}_{i+1})$ is maximal, i.e.,:

$$i = \arg\max_j h(\mathbf{y}_j, \mathbf{y}_{j+1}) \; 1 \le j < M \; . \tag{18}$$

Let the iteration index be set to *t*=1, and denote $\lambda_1^{(t)} = \lambda_i$ and $\lambda_4^{(t)} = \lambda_{i+1}$. We now define hyperparameter search space at iteration *t* as:

$$\boldsymbol{\lambda}^{(t)} := \left[\lambda_1^{(t)} \; \lambda_2^{(t)} \; \lambda_3^{(t)} \; \lambda_4^{(t)}\right] \tag{19}$$

where $\lambda_2^{(t)} = \left(2\times\lambda_1^{(t)} + \lambda_4^{(t)}\right)/3$ and $\lambda_3^{(t)} = \left(\lambda_1^{(t)} + 2\times\lambda_4^{(t)}\right)/3$. We the use the selected SC algorithm to generate pseudo labels $\mathbf{y}_2^{(t)}$ and $\mathbf{y}_3^{(t)}$, [1] and estimate $h\left(\mathbf{y}_1^{(t)},\mathbf{y}_2^{(t)}\right)$, $h\left(\mathbf{y}_2^{(t)},\mathbf{y}_3^{(t)}\right)$ and $h\left(\mathbf{y}_3^{(t)},\mathbf{y}_4^{(t)}\right)$. The subinterval $\left[\lambda_1^{(t+1)},\lambda_4^{(t+1)}\right]$ is refined according to the following rules:

$$
\begin{array}{l}
\text{if } h\left(\mathbf{y}_1^{(t)},\mathbf{y}_2^{(t)}\right)\geq h\left(\mathbf{y}_2^{(t)},\mathbf{y}_3^{(t)}\right) \text{ and } h\left(\mathbf{y}_1^{(t)},\mathbf{y}_2^{(t)}\right)\geq h\left(\mathbf{y}_3^{(t)},\mathbf{y}_4^{(t)}\right)\\
\qquad \lambda_1^{(t+1)} \leftarrow \lambda_1^{(t)},\ \lambda_4^{(t+1)} \leftarrow \lambda_2^{(t)}\\
\text{elseif } h\left(\mathbf{y}_2^{(t)},\mathbf{y}_3^{(t)}\right)\geq h\left(\mathbf{y}_1^{(t)},\mathbf{y}_2^{(t)}\right) \text{ and } h\left(\mathbf{y}_2^{(t)},\mathbf{y}_3^{(t)}\right)\geq h\left(\mathbf{y}_3^{(t)},\mathbf{y}_4^{(t)}\right)\\
\qquad \lambda_1^{(t+1)} \leftarrow \lambda_2^{(t)},\ \lambda_4^{(t+1)} \leftarrow \lambda_3^{(t)}\\
\text{else}\\
\qquad \lambda_1^{(t+1)} \leftarrow \lambda_3^{(t)},\ \lambda_4^{(t+1)} \leftarrow \lambda_4^{(t)}\\
\text{end}
\end{array}
\quad . \tag{20}
$$

We then increment the iteration index to $t \leftarrow t+1$ and repeat the process. HPO stops when the relative error criterion is satisfied:

$$\frac{\lambda_4^{(t+1)} - \lambda_1^{(t+1)}}{\lambda_1^{(t)}} \leq \varepsilon \tag{21}$$

where $\varepsilon$ is a predefined constant. In our experiments, reported in section 4, we set $\varepsilon$=0.001. After the stopping criterion is met, the optimal hypeparameter value is obatined as:

$$\lambda^* = \frac{\lambda_4^{(t+1)} + \lambda_1^{(t+1)}}{2} \tag{22}$$

The experiments presented in section 4 brought to light an important issue related to the selection of the hyperparameter space $\boldsymbol{\lambda} := \left[\lambda_1,...,\lambda_M\right]$. If two neighboring hyperparameters $\lambda_i$ and $\lambda_{i+1}$ are set "too close" to each other, the corresponding metric $h\left(\mathbf{y}_i,\mathbf{y}_{i+1}\right)$ (e.g., ACC or NMI) may exhibit a high value. However, the actual performance based on the true labels, $h\left(\mathbf{y}_i,\mathbf{y}^*\right)$, can be quite poor. This scenario is illustrated in Figure 2, where both ACC and NMI, as functions of $\boldsymbol{\lambda}:=[\lambda_1,...,\lambda_M]$, are estimated by the oracle version of the LSR SC algorithm (left) and pseudo-labels based version of the LSR SC algorithm (right). We summarize our approach to HPO in SC in Algorithm 2.

---

[1] Instead of $\boldsymbol{\lambda}^{(t)} := \left[\lambda_1^{(t)}\ \lambda_2^{(t)}\ \lambda_3^{(t)}\ \lambda_4^{(t)}\right]$ we could use $\boldsymbol{\lambda}^{(t)} := \left[\lambda_1^{(t)}\ \lambda_2^{(t)}\ \lambda_3^{(t)}\right]$, where $\lambda_1^{(t)} = \lambda_i$, $\lambda_3^{(t)} = \lambda_{i+1}$ and $\lambda_2^{(t)} = \left(\lambda_1^{(t)} + \lambda_3^{(t)}\right)/2$. That reduces computational complexity of proposed approach to HPO and affects clustering performance minimally.

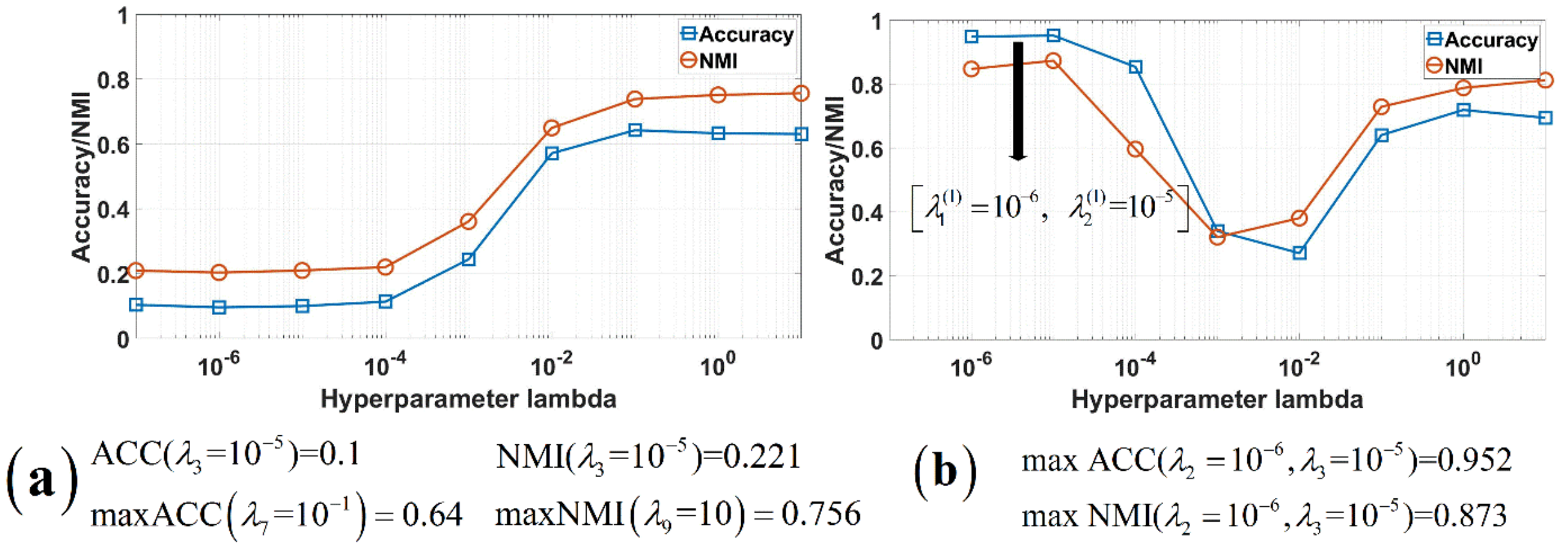

**Figure 2**. Illustration of possible problem in proposed approach to label-free HPO for LSR SC algorithm [17], on COIL20 dataset. Hyperparameter grid values $\lambda_1=10^{-7}$ and $\lambda_2=10^{-6}$ are set too close. (**a**) Accuracy and NMI at $\boldsymbol{\lambda}:=\{10^{-7},10^{-6},10^{-5},10^{-4},10^{-3},10^{-2},10^{-1},1,10\}$. The value 0.6519 of maximal accuracy computed with external (hard) labels occurs at $\lambda_7=0.1$. As for NMI, the maximal value 0.7733 occurs at $\lambda_9=10$. (**b**) Accuracy and NMI computed between pseudo-labels generated by the same grid values of λ as previously. Maximal values occur between pseudo-labels generated by $\lambda_2=10^{-6}$ and $\lambda_3=10^{-5}$. Consequently, new iteration 1 begins on the interval with borders $\lambda_1^{(1)}=\lambda_2=10^{-6}$ and $\lambda_2^{(1)}=\lambda_3=10^{-5}$. That is incorrect.

**Algorithm 2.** Proposed approach to a HPO for a single hyperparameter SC algorithm (e.g., LSR SC equation (6), [17]).

**Input**: Feature (data) matrix **X**, initial hyperparameter space $\boldsymbol{\lambda}:=[\lambda_1,...,\lambda_M]$, stopping criterion $\varepsilon=0.001$.

1: Locate the subinterval $[\lambda_i,\lambda_{i+1}]$ where $h(\mathbf{y}_i,\mathbf{y}_{i+1})$ is maximal, i.e.:

$$i=\arg\max_j h(\mathbf{y}_j,\mathbf{y}_{j+1})\ 1\le j<M \quad (18).$$

2: Set $t$=1, and denote $\lambda_1^{(t)}=\lambda_i$, $\lambda_4^{(t)}=\lambda_{i+1}$, $\mathbf{y}_1^{(t)}=\mathbf{y}_i$ and $\mathbf{y}_4^{(t)}=\mathbf{y}_{i+1}$.

3: Set $\lambda_2^{(t)} = \left(2\times\lambda_1^{(t)} + \lambda_4^{(t)}\right)/3$ and $\lambda_3^{(t)} = \left(\lambda_1^{(t)} + 2\times\lambda_4^{(t)}\right)/3$.

4: Use the selected SC algorithm to generate pseudo labels $\mathbf{y}_2^{(t)}$ and $\mathbf{y}_3^{(t)}$.

5: Estimate $h\left(\mathbf{y}_1^{(t)},\mathbf{y}_2^{(t)}\right)$, $h\left(\mathbf{y}_2^{(t)},\mathbf{y}_3^{(t)}\right)$ and $h\left(\mathbf{y}_3^{(t)},\mathbf{y}_4^{(t)}\right)$.

6: Use equation (20) to select borders of the next subinterval $\lambda_1^{(t+1)}$ and $\lambda_4^{(t+1)}$.

7: If the relative error criterion (13) is met, proceed to step 13.

8: Compute $\lambda_2^{(t+1)} = \left(2\times\lambda_1^{(t+1)} + \lambda_4^{t+1}\right)/3$, and $\lambda_3^{(t+1)} = \left(\lambda_1^{(t+1)} + 2\times\lambda_4^{t+1}\right)/3$.

9: Use selected SC algorithm to generate pseudo labels $\mathbf{y}_1^{(t+1)}$, $\mathbf{y}_2^{(t+1)}$, $\mathbf{y}_3^{(t+1)}$ and $\mathbf{y}_4^{(t+1)}$.

10: Estimate $h\left(\mathbf{y}_1^{(t+1)},\mathbf{y}_2^{(t+1)}\right)$, $h\left(\mathbf{y}_2^{(t+1)},\mathbf{y}_3^{(t+1)}\right)$ and $h\left(\mathbf{y}_3^{(t+1)},\mathbf{y}_4^{(t+1)}\right)$.

11: Increment $t \leftarrow t+1$.

12: Return to step 6.

13: Compute optimal hyperparameter value $\lambda^*$ using equation (22).

**Output**: The optimal hyperparameter value $\lambda^*$.

*3.2. Extension of proposed approach to HPO for SC algorithm with two hyperparameters*

We now extend our approach to HPO for SC algorithms with two hyperparameters, such as kernel LSR SC (7) (assuming Gaussian kernel), S0L0 LRSSC algorithm [16], graph filtering LSR SC [68] (also presented in Algorithm 1), and the MCLME algorithm [72]. Let us assume that the hyperparameter space is defined by $\boldsymbol{\lambda} := \{\lambda_1, ..., \lambda_M\}$ and $\boldsymbol{\tau} := \{\tau_1, ..., \tau_L\}$. First, we preset the hyperparameter $\tau$ to a value $\tau = \tau_{L/2}$ (other options may also be considered). We then apply Algorithm 1 with the predefined hyperparameter space $\boldsymbol{\lambda}$ to compute $\lambda^*$. Afterward, we fix $\lambda$ at the obtained value $\lambda = \lambda^*$, and apply Algorithm 1 with the predefined hyperparameter space $\boldsymbol{\tau}$ to compute $\tau^*$. Described approach is based on implicit assumptions on $h_{\tau_{L/2}}\left(\mathbf{y}_i, \mathbf{y}_{i+1}\right)$ and $h_{\lambda^*}(\mathbf{y}_i,\mathbf{y}_{i+1})$, analogous to equations (16) and (17). Specifically, we assume that $h_{\tau_{L/2}}(\mathbf{y}_i,\mathbf{y}_{i+1})$ is a monotonic function of $\tau_i$ and $\tau_{i+1}$, and $h_{\lambda^*}(\mathbf{y}_i,\mathbf{y}_{i+1})$ is a monotonic function of $\lambda_i$ and $\lambda_{i+1}$. The subscripts $\tau_{L/2}$ and $\lambda^*$ indicate that thet performance indices are computed with these parameters. Described method for HPO of the SC algorithms with two hypeparameters can be straightforwardly extended to HPO of the SC algorithms that depend on three or more hyperparameters.

*3.3. Extension of proposed approach to HPO of SC algorithm for out-of-sample data*

Many SC algorithms are unable to cluster out-of-sample (also known as unseen or test) data [14-30]. This limitation applies to deep SC algorithms [32-35, 63] and tensor-based SC algorithms [71] as well. In other words, to cluster an unseen data point, the algorithm must be re-run again on the entire, including the new data point. This requirement significantly reduces the scalability and applicability of these algorithms in large-scale or online clustering problems. Here, we address the problem of clustering out-of-sample data by formulating it as a minimization problem based on the point-to-a-subspace distance criterion. Using the partitions obtained by the selected algorithm on the in-sample dataset, we estimate the subspace bases [37]:

$$\left\{ \mathbf{X}_c \leftarrow \mathbf{X}_c - \left[ \underbrace{\bar{\mathbf{x}}_c \ldots \bar{\mathbf{x}}_c}_{N_c \text{ times}} \right] \right\}_{c=1}^{C} \tag{23}$$

where $\bar{\mathbf{x}}_c = \frac{1}{N_c}\sum_{n=1}^{N_c}\mathbf{X}_c(n)$, $\bigcup_{c=1}^{C}\mathbf{X}_c = \mathbf{X}$, $\sum_{c=1}^{C} N_c = N$, and $C$ is the number of clusters. From $\left\{ \mathbf{X}_c = \mathbf{U}_c \mathbf{\Sigma}_c (\mathbf{V}_c)^{\mathrm{T}} \right\}_{c=1}^{C}$ we estimate the orthonormal bases by extracting the first $d$ left singular vectors of each partition, i.e. $\left\{ \mathbf{U}_c \in \mathbb{R}^{D\times d} \right\}_{c=1}^{C}$ [37]. For a test data point $\mathbf{x}$, we compute the point-to-subspace distances $\left\{ d_c = \left\| \tilde{\mathbf{x}}_c - \mathbf{U}_c (\mathbf{U}_c)^{\mathrm{T}} \tilde{\mathbf{x}}_c \right\|_2 \right\}$ as follows:

$$c = \underset{c\in\{1,\ldots,C\}}{\arg\min}\, d_c \tag{24}$$

where $\tilde{\mathbf{x}}_c = \mathbf{x} - \overline{\mathbf{x}}_c$. We assign the label $\{c\}_{c=1}^{C}$ to the test data point:

$$\left[\pi(\mathbf{x})\right]_c = \begin{cases} 1, \text{ if } c = \underset{c\in\{1,\dots,C\}}{\arg\min}\, d_c \\ 0, \text{ otherwise.} \end{cases} \quad (25)$$

*3.4. Formulation of proposed approach to HPO for out-of-sample extension of the kernel LSR SC algorithm*

To apply the point-to-subspace distance criterion for clustering out-of-sample data in reproducible kernel Hilbert space (RKHS), the approach presented in section 3.3 has to be slightly adapted. First, we need to estimate the subspace bases in the kernel-induced RKHS. Let $\Phi(\mathbf{X}) \triangleq \left[\phi(\mathbf{x}_1),\dots,\phi(\mathbf{x}_N)\right] \in \mathbb{R}^{F\times N}$ represent the in-sample data X in the $F$-dimensional feature space. Wes assume $\Phi(\mathbf{X})$ has zero mean. If $\Psi(\mathbf{X})$ represents the non-zero mean version of the mapped in-sample data, then $\Phi(\mathbf{X})$ is computed as:

$$\Phi(\mathbf{X}) = \Psi(\mathbf{X})(\mathbf{I}_N - \mathbf{E}_N) \quad (26)$$

with $\mathbf{I}_N$ being the $N\times N$ identity matrix, and $\mathbf{E}_N \triangleq \frac{1}{N}\mathbf{1}_N\mathbf{1}_N^{\mathrm{T}}$ is an $N\times N$ matrix with all elements equal to $1/N$. Let $\mathbf{K} \triangleq \Phi(\mathbf{X})^{\mathrm{T}}\Phi(\mathbf{X}) \in \mathbb{R}^{N\times N}$ be the kernel (Gram) matrix of the in-sample data, with rank $R$. Since $\Phi(\mathbf{X})$ is centered, it holds that $R \le N-1$. If $\mathcal{K} \triangleq \Psi(\mathbf{X})^{\mathrm{T}}\Psi(\mathbf{X})$ is the non-centered kernel matrix, the centered kernel matrix is obtained as:

$$\mathbf{K} = (\mathbf{I}_N - \mathbf{E}_N)\mathcal{K}(\mathbf{I}_N - \mathbf{E}_N) \ . \quad (27)$$

Let the kernel vector associated with any test point $\mathbf{x} \in \mathbb{R}^D$ be defined as:

$$k(\mathbf{x}) \triangleq \Phi(\mathbf{X})^{\mathrm{T}}\phi(\mathbf{x}) \in \mathbb{R}^N \ . \quad (28)$$

Let $\kappa(\mathbf{x})$ be non-centered kernel vector. We obtain the centered one as:

$$k(\mathbf{x}) = (\mathbf{I}_N - \mathbf{E}_N)\left[\kappa(\mathbf{x}) - \tfrac{1}{N}\mathcal{K}\mathbf{1}_N\right]. \quad (29)$$

Let $\mathcal{P}$ be the $R$-dimensional subspace of the feature space spanned by $\Phi(\mathbf{X})$. Let $\phi_{\mathbf{w}}(\mathbf{x})$ be the projection of $\phi(\mathbf{x})$ on the 1-D vector space spanned by $\mathbf{w} \in \mathbb{R}^F$. We restrict $\mathbf{w}$ to be in $\mathcal{P}$, i.e.

$\mathbf{w}=\Phi(\mathbf{X})\boldsymbol{\alpha}$ for some $\boldsymbol{\alpha} \in \mathbb{R}^N$. Let the eigenvalue decomposition of $\mathbf{K}$ be $\mathbf{K} = \mathbf{U}\boldsymbol{\Lambda}\mathbf{U}^{\mathrm{T}}$, where $\mathbf{U} \in \mathbb{R}^{N\times R}$ and $\boldsymbol{\Lambda} = \mathrm{diag}(\lambda_1,...,\lambda_R) \in \mathbb{R}^{R\times R}$. The orthonormal basis of $\mathcal{P}$ is obtained as [74]:

$$\boldsymbol{\Pi} \triangleq \Phi(\mathbf{X})\mathbf{U}\boldsymbol{\Lambda}^{-1/2} = [\boldsymbol{\pi}_1,...,\boldsymbol{\pi}_R] \in \mathbb{R}^{F\times R} \,. \quad (30)$$

The coordinates of the mapped training data $\Phi(\mathbf{X})$ in the Cartesian coordinate system spanned by $\boldsymbol{\Pi}$ are:

$$\mathbf{Y} = \boldsymbol{\Lambda}^{-1/2}\mathbf{U}^{\mathrm{T}}\mathbf{K} = \boldsymbol{\Lambda}^{1/2}\mathbf{U}^{\mathrm{T}} \in \mathbb{R}^{R\times N} \,. \quad (31)$$

Similarly, the coordinates of the projection of any $\mathbf{x} \in \mathbb{R}^D$ onto $\mathcal{P}$ in the same coordinate system are obtained as:

$$\mathbf{y} = \boldsymbol{\Pi}^{\mathrm{T}}\phi(\mathbf{x}) = \boldsymbol{\Lambda}^{-1/2}\mathbf{U}^{\mathrm{T}}k(\mathbf{x}) \,. \quad (32)$$

At this point, we can formally replace $\mathbf{X}$ with $\mathbf{Y}$ in (23) and $\mathbf{x}$ with $\mathbf{y}$ in (24) and (25) to assign the cluster label to a test data point $\mathbf{y}$ in RKHS. As mentioned in (7), the Gaussian kernel LRS algorithm introduces one hyperparameter, the variance $\sigma^2$. However, clustering out-of-sample data introduces in another hyperparameter: the rank $R$ of the centered kernel matrix $\mathbf{K}$ formed from the in-sample data. To simplify the experiments, we set $R=N-1$, in the kernel LSR algorithm.

### *3.5 Interpretable LFSG SC*

As with many other complex prediction models, SC algorithms generally suffer from the issue of interpretability. In our approach to LFSG SC, we propose a method to interpret or explain the results obtained by the LFSG SC algorithm. Notably, this proposed approach is also applicable to the oracle versions of the corresponding SC algorithm. Our method relies on the visualization of subspace bases estimated from the in-sample data, using the clustering partitions produced by the selected SC algorithm. Let the SVD of each in-sample cluster partition be represented as:

$$\left\{\mathbf{X}_c = \mathbf{U}_c\boldsymbol{\Sigma}_c(\mathbf{V}_c)^{\mathrm{T}}\right\}_{c=1}^{C} \,. \quad (33)$$

We aim to visualize each partition. To achieve this, we compute the representative of each cluster as:

$$\left\{ \mathbf{a}_c \in \mathbb{R}^{D\times 1} = \sum_{j=1}^{d} \mathbf{U}_c\left(:,j\right)\mathbf{\Sigma}_c\left(j,j\right) \right\}_{c=1}^{C} . \tag{34}$$

In equation (34), $d$ represents the subspace dimension, which is known *a priori*. For example, face images of each subject in the EYaleB dataset lie approximately in a subspace of $d$=9 dimensions [75], while handwritten digits such as those in the MNIST dataset, appriximately lie in a subspace of $d$=12 dimensions [76]. Since in SC algorithms typically vectorize image data samples, we need to matricize the representative vectors to recover their original 2D form. Let us assume the dimensions of the imaging data are $D_x$ and $D_y$, where $D=D_x \times D_y$. Using MATLAB notation, we reconstruct the images of the cluster representatives as:

$$\left\{ \mathbf{A}_c \in \mathbb{R}^{D_x \times D_y} = reshape\left(\mathbf{a}_c, D_x, D_y\right) \right\}_{c=1}^{C} . \tag{35}$$

We demonstrate the proposed visualization method on the USPS dataset [78], which contains images of handwritten numerals organized into $C$=10 clusters. Each grayscale image has dimensions $D_x=D_y=16$. Figure 3 shows the visualization results for correctly labeled data, as well as for data labeled by both the oracle and LFSG versions of the S0L0 LRSSC algorithm [16]. Despite the inherent rotational indeterminacy of the proposed visualization, the numerals corresponding to each group are recognizable, allowing us to explain the decisions of both the oracle LFGS versions of the S0L0 LRSSC algorithm.

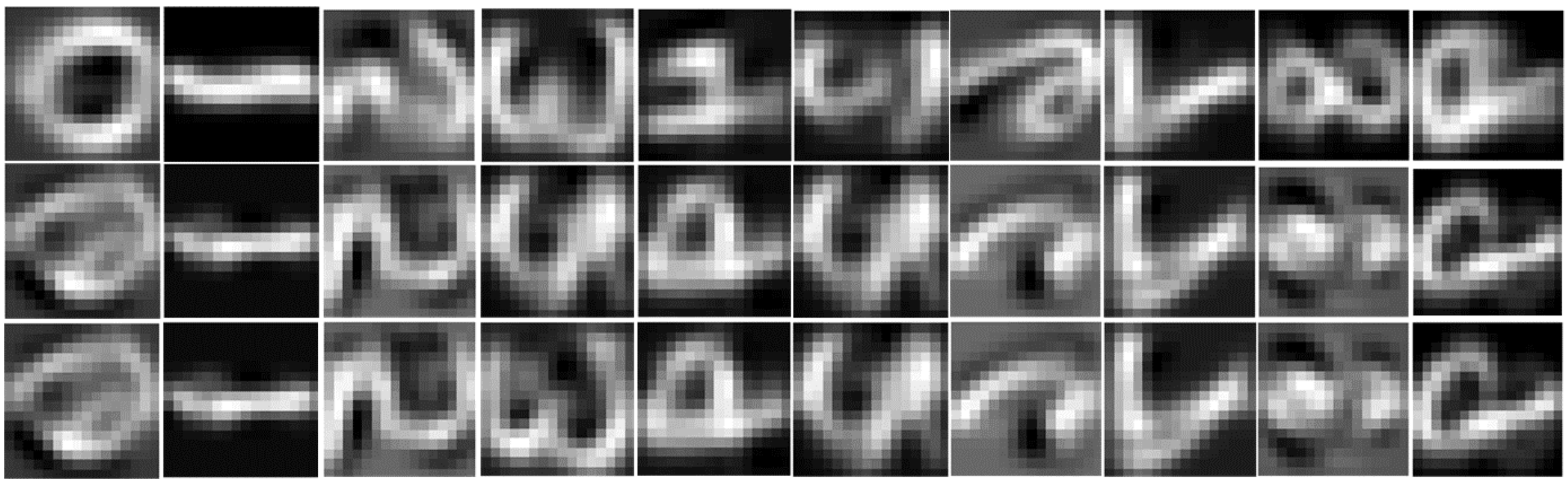

**Figure 3**. Visualization of subspace bases representing the numerals in the USPS dataset [78], from 0 to 9 in order. The bases are estimated from 50 randomly selected data samples per cluster (numeral). Clustering is performed using the S0L0 LRSSC algorithm [16]. From top to bottom: partitions are composed of (1) correctly labeled data, (2) data labeled by the oracle version of the S0L0 LRSSC algorithm, (3) data labeled by the LFSG version of the S0L0 LRSSC algorithm. The clustering accuracy of both versions is approximately 85%.

## 4.0 Experiments

### *4.1. Single-view SC algorithms*

We evaluate the proposed approach to HPO on LSR SC algorithm (6), [17], graph-filtering version of LSR SC algorithm in Algorithm 1, kernel LSR SC algorithm (7), [17], sparse SC (SSC) algorithm [15], and S0L0 low rank sparse SC (LRSSC) algorithm [16]. For this evaluation we use six datasets: MNIST [77], USPS [78], EYaleB [79], ORL [80], COIL20, and COIL100 [81]. For each algorithm, we compute accuracy and normalized mutual information (NMI) to select the optimal hyperparameter value over the specified search space. This is done using both the true labels (the oracle) and pseudo-labels, following the label-free self-guided (LFSG) approach. After determining the optimal hypeparameter value, we compute accuracy, NMI and F1-score using the true labels to assess clustering performance. To evaluate the robustness of the algorithms, we randomly split each dataset into in-sample and out-of-sample partitions and repeat the clustering process 25 times. This allows us to statistically compare the oracle and  LFSG versions of each selected SC algorithm using the Wilcoxon rank-sum test implemented by MATLAB command `ranksum`. The null hypothesis of the test assumes that the data come from continuous distributions with equal medians at a 5% significance level. A p-value greater than 0.05 indicates acception of the null hypothesis. The main information on datasets used in the experiments are summarized in Table 1.

**Table 1**
Main information on datasets used in the experiments with single-view SC algorithms

| Dataset | #Sample | #Feature | #Cluster | #in-sample/out-of-sample per group |
|---|---|---|---|---|
| MNIST | 10000 | 28×28 | 10 | 50/50 |
| USPS | 7291 | 16×16 | 10 | 50/50 |
| EYaleB | 2432 | 48×42 | 38 | 43/21 |
| ORL | 400 | 32×32 | 40 | 7/3 |
| COIL20 | 1440 | 32×32 | 20 | 26/26 |
| COIL100 | 7200 | 32×32 | 100 | 26/26 |

*4.1.1. Least squares regression SC algorithm*

Table 2 presents the clustering performance of the oracle and LFSG versions of the LSR SC algorithm. In both cases, the hyperparameter λ was selected based on either accuracy (ACC) or NMI. For each scenario, we report the performance metrics on both in-sample and out-of-sample data as mean±standard deviation over 25 runs. Additionally, p-values are provided at the 95% confidence interval for instances when the statistical difference between the oracle and LFSG versions was insignificant. These findings suggests that the proposed LFSG LSR SC algorithm performs comparably to its oracle counterpart. For further details on the hyperparameter λ search space for each dataset, we direct interested readers to the publicly available code. As shown in Table 2, the LFSG version on in-sample data typically has up to 4% lower clustering performance to the oracle version, while on out-of-sample data, the performance difference is typically less than 1%. This suggests that the partitions generated by the LFSG version on in-sample data are sufficiently robust to estimate accurate subspace bases for clustering out-of-sample (test) data. Additionally, in cases of MNIST and USPS datasets, the differences in performance and hyperparameter estimates between the two versions are statistically insignificant.

**Table 2**
The clustering performance of the oracle and LFSG versions of the LSR SC algorithm. In both versions, the hyperparameter λ is selected based on either accuracy (ACC) or NMI. For each scenario, we report performance metrics achieved on in-sample and out-of-sample data. Bold font indicates cases where the difference between the oracle and LFSG versions is statistically insignificant.

| **Dataset** | | **ACC [%]**<br>in-sample<br>out-of-sample | **NMI [%]**<br>in-sample<br>out-of-sample | **F1 score[%]**<br>in-sample<br>out-of-sample | **Hyperparameter** |
|---|---|---|---|---|---|
| ORL | ORACLE-ACC | 69.34±2.57<br>68.75±2.46 | 83.94±1.27<br>86.41±1.63 | 56.57±2.96<br>47.87±5.03 | λ=0.494±1.703 |
| | LFSG-ACC | 66.92±3.31 | 82.55±1.86 | 53.62±4.10 | λ=0.488±1.097 |
| | | 67.28±3.01 | 85.85±1.51 | 46.34±4.31 | |
| | ORACLE-NMI | 68.18±1.98 | 83.63±1.11 | 55.73±2.47 | λ=0.142±0.157 |
| | | 67.42±2.77 | 86.50±1.25 | 47.90±3.85 | |
| | LFSG-NMI | 66.82±2.92 | 82.57±1.64 | 53.68±3.45 | λ=0.336±0.347 |
| | | 66.77±3.65 | 85.72±1.40 | 45.97±4.27 | |
| EYaleB | ORACLE-ACC | 66.56±1.78 | 74.73±1.01 | 55.67±1.69 | λ=0.1±0.0 |
| | | 70.24±3.19 | 80.03±1.89 | 61.48±2.98 | |
| | LFSG-ACC | 67.16±5.20 | 74.67±3.14 | 55.07±4.64 | λ=0.104±0.161 |
| | | 69.87±5.60 | 79.69±3.34 | 60.56±2.35 | |
| | | **p_in= 0.058** | **p_in=0.237** | **p_in=0.561** | |
| | | **p_out = 0.467** | **p_out=0.628** | **p_out=0.938** | |
| | ORACLE-NMI | 66.96±3.13 | 75.10±1.89 | 56.28±2.30 | λ=0.1±0.0 |
| | | 70.15±1.96 | 80.12±1.24 | 61.25±2.03 | |
| | LFSG-NMI | 65.68±5.52 | 74.22±3.29 | 54.13±4.42 | λ=0.140±0.170 |
| | | 68.54±5.74 | 79.04±3.33 | 59.44±5.05 | |
| | | **p_in=0.816** | **p_in=0.831** | **p_in=0.229** | |
| | | **p_out=0.838** | **p_out=0.614** | **p_out=0.522** | |
| MNIST | ORACLE-ACC | 57.79±3.92 | 56.40±5.58 | 46.61±2.86 | λ=4.40±3.31 |
| | | 57.55±3.81 | 57.14±2.48 | 46.88±2.72 | |
| | LFSG-ACC | 56.10±4.12 | 55.22±3.26 | 45.41±3.09 | λ=4.81±2.20 |
| | | 56.65±4.32 | 57.16±2.36 | 46.73±2.74 | **p=0.654** |
| | | **p = 0.140** | **p_in =0.174** | **p_in = 0.088** | |
| | | **p =0.600** | **p_out = 0.985** | **p_out =0.938** | |
| | ORACLE-NMI | 57.06±3.60 | 55.94±2.26 | 46.18±2.53 | λ=5.54±2.79 |
| | | 57.42±4.65 | 58.21±3.07 | 47.47±3.84 | |
| | LFSG-NMI | 54.78±4.13 | 53.96±2.48 | 44.09±2.77 | λ=6.07±2.55 |
| | | 56.74±4.53 | 57.26±3.04 | 46.47±3.65 | **p=0.907** |
| | | **p_out = 0.705** | **p_out= 0.260** | **p_out = 0.415** | |
| USPS | ORACLE-ACC | 75.84±3.89 | 70.90±2.19 | 64.17±2.81 | λ=4.48±3.52 |
| | | 76.30±3.55 | 71.86±1.75 | 65.01±2.44 | |
| | LFSG-ACC | 74.66±4.05 | 69.68±2.15 | 62.94±2.82 | λ=3.96±1.04 |
| | | 75.00±4.48 | 71.06±2.05 | 63.83±2.84 | **p=0.585** |
| | | **p_in=0.322** | **p_in=0.05** | **p_in=0.168** | |
| | | **p_out=0.256** | **p_out=0.181** | **p_out=0.121** | |
| | ORACLE-NMI | 76.74±3.15 | 70.57±1.59 | 64.27±2.37 | λ=2.84±3.25 |
| | | 74.20±3.34 | 70.84±2.01 | 63.77±2.41 | |
| | LFSG-NMI | 73.30±3.87 | 68.60±2.02 | 61.91±2.90 | λ=4.98±2.33 |
| | | 73.12±3.84 | 69.35±2.33 | 61.98±2.73 | |
| | | **p_out=0.204** | | | |
| COIL20 | ORACLE-ACC | 68.53±1.97 | 77.93±1.19 | 61.55±1.94 | λ=3.45±3.95 |
| | | 64.02±2.11 | 74.77±1.95 | 58.55±3.04 | |
| | LFSG-ACC | 56.75±20.94 | 66.45±20.75 | 58.36±2.76 | λ=1.07±1.18 |
| | | 56.07±19.39 | 67.13±17.85 | 57.29±2.82 | |
| | | **p_out=0.497** | **p_out=0.313** | **p_out=0.151** | |
| | ORACLE-NMI | 69.44±1.70 | 78.03±1.25 | 61.35±2.00 | λ=5.22±3.93 |
| | | 65.68±2.84 | 76.28±1.67 | 58.24±2.95 | |
| | LFSG-NMI | 65.86±2.96 | 76.14±1.64 | 57.45±3.00 | λ=3.13±3.09 |
| | | 65.06±2.88 | 75.57±1.62 | 56.63±3.14 | |
| | | **p_out=0.426** | **p_out=0.269** | | |
| COIL100 | ORACLE-ACC | 51.88±0.79<br>49.62±1.28 | 76.52±0.39<br>75.52±0.68 | 44.66±0.76<br>42.42±1.31 | λ=5.49±3.78 |
| | LFSG-ACC | 49.45±1.58<br>48.60±1.28 | 74.52±0.90<br>74.27±0.81 | 41.31±1.62<br>40.50±1.54 | λ=1.04±1.70 |
| | ORACLE-NMI | 51.49±1.02<br>49.31±1.48 | 76.33±0.40<br>75.57±0.48 | 44.34±0.82<br>42.56±1.25 | λ=7.04±3.05 |
| | LFSG-NMI | 49.76±1.44<br>48.90±1.50<br>**p_out=0.509** | 74.67±0.75<br>74.51±0.59 | 41.73±1.53<br>40.99±1.33 | λ=0.901±0.419 |

*4.1.2. Kernel least squares regression SC algorithm*

For the kernel LSR SC algorithm (7), we selected the Gaussian kernel, introducing variance $\sigma^2$ as an additional hyperparameter alongside λ. Table 3 presents the clustering performance of both the oracle and LFSG versions of the kernel LSR SC algorithm, with presentation details similar to those in Table 2. As observed in Table 3, the oracle version typically outperforms the LFSG version by 3-4%. However, in several cases, the performance difference on in-sample data is statistically insignificant. Notably, for out-of-sample data, the LFSG version achieves significantly better performance on all datasets. This suggests that the partitions generated by the LFSG version on in-sample data are more robust for estimating accurate subspace bases for clustering out-of-sample (test) data. It can be argued that the oracle version could also produce more robust partitions if a denser hyperparameter search space were employed. However, this would significantly increase computational complexity.

**Table 3**

The clustering performance of the oracle and LFSG versions of the kernel LSR SC algorithm. In each version, the hyperparameters λ and $\sigma^2$ were selected based on either ACC or NMI. Performance metrics were calculated for both in-sample and out-of-sample data. Bold font indicates cases where the difference between the oracle and LFSG versions is statistically insignificant.

| **Dataset** | | **ACC [%]**<br>in-sample<br>out-of-sample | **NMI [%]**<br>in-sample<br>out-of-sample | **F_score[%]**<br>in-sample<br>out-of-sample | **Hyperparameters** |
|---|---|---|---|---|---|
| ORL | ORACLE-ACC | 67.83±3.07<br>47.03±11.29 | 83.56±1.76<br>73.08±6.86 | 55.49±3.86<br>21.11±12.21 | λ=0.0012±0.0018<br>$\sigma^2$=564±1977 |
| | LFSG-ACC | 65.56±4.49<br>66.47±3.97 | 81.66±2.59<br>85.55±1.92 | 51.67±5.35<br>45.48±5.42 | λ=0.0135±0.0267<br>$\sigma^2$=1037±2002 |
| | ORACLE-NMI | 67.60±2.66<br>46.47±11.30 | 83.26±1.63<br>73.28±7.28 | 54.85±3.32<br>21.41±11.80 | λ=0.0051±0.028<br>$\sigma^2$=236±273.68 |
| | LFSG-NMI | 65.27±2.41<br>66.87±2.54 | 81.67±1.55<br>85.62±1.40 | 51.56±3.08<br>45.81±4.34 | λ=0.0127±0.0294<br>**p=0.059**<br>$\sigma^2$=349.62±222.40 |
| EYaleB | ORACLE-ACC | 66.69±2.43<br>34.41±15.98 | 74.58±1.52<br>45.90±14.85 | 55.17±1.86<br>19.19±13.67 | $\sigma^2$=100±0.0<br>λ=0.001±0.0 |
| | LFSG-ACC | 55.30±13.48<br>58.78±13.22 | 66.56±9.33<br>72.55±8.82 | 44.27±12.17<br>50.06±12.36<br>**p_in=0.189** | $\sigma^2$=1796±2791<br>λ=0.00062±0.00018 |
| | ORACLE-NMI | 67.09±2.83<br>38.83±15.73 | 74.78±1.90<br>49.97±14.52 | 55.24±2.79<br>22.34±13.75 | $\sigma^2$=100±0.00<br>λ=0.001±0.0 |
| | LFSG-NMI | 64.071±3.72<br>67.47±3.55 | 72.58±2.72<br>78.54±2.22 | 52.04±3.66<br>58.81±3.51 | $\sigma^2$=327.91±196.72<br>λ=0.00065±0.00018 |
| MNIST | ORACLE-ACC | 57.14±4.42<br>40.99±19.19 | 54.51±3.74<br>29.62±17.52 | 45.43±4.19<br>27.20±12.28 | $\sigma^2$=12269±24232<br>λ=0.0943±0.1827 |
| | LFSG-ACC | 55.27±3.81<br>55.93±3.41<br>**p_in = 0.064** | 52.11±3.40<br>54.80±3.09 | 45.33±3.27<br>44.97±3.11 | $\sigma^2$=26505±29472<br>**p=0.05**<br>λ=0.1493±0.1775 |
| | ORACLE-NMI | 55.81±4.42<br>29.19±5.18 | 55.72±2.65<br>19.64±7.14 | 42.26±2.59<br>19.03±4.21 | $\sigma^2$=736±946<br>λ=0.0073±0.0094 |
| | LFSG-NMI | 54.50±2.87<br>56.17±3.18<br>**p_in = 0.286** | 50.87±2.56<br>54.42±3.00 | 41.83±3.37<br>45.16±3.50 | $\sigma^2$=49928±33692<br>λ=0.1191±0.1697 |
| USPS | ORACLE-ACC | 74.98±4.49<br>53.26±19.03 | 69.54±2.64<br>40.61±21.31 | 63.16±3.37<br>38.32±18.14 | $\sigma^2$=15286±30201<br>λ=0.0258±0.0362 |
| | LFSG-ACC | 72.74±4.47<br>73.06±4.65<br>**p_in= 0.123** | 66.45±3.38<br>67.38±3.78 | 60.17±3.85<br>61.03±4.54 | $\sigma^2$=30509±35294<br>**p= 0.124**<br>λ=0.1358±0.1302 |
| | ORACLE-NMI | 73.70±4.24<br>51.41±20.85 | 71.08±2.10<br>40.77±25.61 | 63.86±2.96<br>38.85±20.87 | $\sigma^2$=4398±19942<br>λ=0.0559±0.1366 |
| | LFSG-NMI | 69.90±3.48<br>69.77±2.59 | 65.11±4.02<br>65.35±3.89 | 58.17±4.15<br>58.04±3.70<br>**p_out= 0.057** | $\sigma^2$=28883±35955<br>λ=0.0611±0.0946 |
| COIL20 | ORACLE-ACC | 65.87±3.58<br>62.12±9.13 | 76.44±2.07<br>73.00±1.57 | 60.32±3.01<br>54.08±10.69 | $\sigma^2$=10586±21067<br>λ=0.0827±0.1607 |
| | LFSG-ACC | 66.13±3.07<br>67.00±3.02<br>**p_in=0.907** | 76.47±1.77<br>77.11±1.66<br>**p_in= 0.985**<br>**p_out=0.046** | 60.19±2.90<br>59.56±2.86<br>**p_in=0.877** | $\sigma^2$=6986±18328<br>**p_acc=0.062**<br>λ=0.103±0.1493<br>**p_acc=0.228** |
| | ORACLE-NMI | 66.60±2.59<br>63.85±6.61 | 77.35±1.38<br>74.99±6.13 | 61.49±2.65<br>57.07±8.13 | $\sigma^2$=11720±20584<br>λ=0.0919±0.1831 |
| | LFSG-NMI | 65.99±3.49<br>66.90±3.41<br>**p_in=0.669**<br>**p_out=0.771** | 76.63±2.03<br>77.52±1.76<br>**p_in= 0.252**<br>**p_out=0.260** | 60.51±3.46<br>60.36±3.03<br>**p_in= 0.393**<br>**p_out=0.200** | $\sigma^2$=8290±18194<br>λ=0.0651±0.0948<br>**p_nmi=0.139** |
| COIL100 | ORACLE-ACC | 50.89±1.02<br>43.94±10.18 | 76.16±0.66<br>70.35±8.75 | 45.09±1.07<br>34.59±11.64 | $\sigma^2$=6700±9968<br>λ=0.057±0.1362 |
| | LFSG-ACC | 48.26±2.37<br>49.58±2.14<br>**p=0.181** | 74.14±1.75<br>75.73±1.33<br>**p=0.044** | 42.15±2.85<br>42.07±2.40 | $\sigma^2$=352.9±342.8<br>λ=0.0131±0.0219<br>**p=0.149** |
| | ORACLE-NMI | 50.06±1.21<br>43.49±10.71 | 76.14±0.43<br>70.04±9.34 | 44.68±0.94<br>34.03±12.58 | $\sigma^2$=7320±9817<br>λ=0.0366±0.1005 |
| | LFSG-NMI | 49.81±2.04<br>51.04±1.90<br>**p_in=0.839** | 75.04±1.28<br>76.47±0.92 | 43.76±2.19<br>43.56±1.80<br>**p_in= 0.0.146** | $\sigma^2$=535.4±326.3<br>λ=0.0327±0.0814<br>**p=0.149** |

*4.1.3. Graph filtering least squares regression SC algorithm*

As shown in Algorithm 1, the graph-filtering LSR SC algorithm introduces an additional hyperparameter, the filter order *k*, alongside the regularization constant λ. Unlike the LSR SC and kernel LSR SC algorithms, we have not yet formulated an out-of-sample extension for the graph-filtering LSR SC algorithm. Therefore, all results presented in Table 4 are based on in-sample-data. As seen in Table 4, the performance difference between the oracle and LFSG versions is typically less than 4%, and sometimes below 2%. For the ORL, COIL20, and partially USPS datasets, there is no statistically significant difference in performance between the two versions. In several instances, there is also no statistically significant difference in hyperparameter estimates obtained by the oracle and LFSG versions.

**Table 4**

The clustering performance of the oracle and LFSG versions of the graph-filtering LSR SC algorithm. In each version, the hyperparameters λ and *k* were selected based on either ACC or NMI. Performance metrics were calculated using in-sample data. Bold fond is used to indicate instances where the difference between the oracle and LFSG versions is statistically insignificant.

| Dataset | | ACC [%] | NMI [%] | F_score[%] | Hyperparameters |
|---|---|---|---|---|---|
| ORL | ORACLE-ACC | 67.56±2.09 | 82.62±1.22 | 56.58±2.14 | λ=0.0067±0.0198<br>k=4.28±1.79 |
| | LFSG-ACC | 66.75±2.28 **p = 0.193** | 82.15±1.44 **p=0.295** | 55.63±2.76 **p=0.237** | λ=0.0140±0.0233 k=3.72±2.23 **p=0.048** |
| | ORACLE-NMI | 66.29±2.56 | 82.05±1.43 | 54.82±2.85 | λ=0.0140±0.0326 k=4.16±2.03 |
| | LFSG-NMI | 65.86±3.06 **p=0.771** | 81.44±1.76 **p= 0.237** | 54.33±3.47 **p=0.801** | λ=0.0184±0.0251 k=4.04±2.78 **p=0.519** |
| EYaleB | ORACLE-ACC | 66.23±2.67 | 73.10±2.03 | 54.52±2.31 | λ=0.0013±0.0018 k=3.52±0.59 |
| | LFSG-ACC | 62.72±4.17 | 67.78±6.16 | 47.72±7.00 | λ=0.0025±0.0025 **p=0.488** k=2.24±0.44 |
| | ORACLE-NMI | 66.10±2.66 | 73.26±1.87 | 54.63±1.86 | λ=0.0014±0.0018 k=3.68±0.56 |
| | LFSG-NMI | 59.73±4.11 | 66.95±4.14 | 46.41±4.91 | λ=0.0018±0.0023 k=2.24±0.44 |
| MNIST | ORACLE-ACC | 56.68±2.09 | 52.76±2.52 | 45.42±2.32 | λ=0.0042±0.138 k=7.12±1.13 |
| | LFSG-ACC | 53.88±2.64 | 49.31±2.05 | 42.32±2.05 | λ=0.259±0.166 k=8.2±1.38 |
| | ORACLE-NMI | 55.62±3.99 | 54.75±2.13 | 46.44±2.62 | λ=0.00096±0.00018 k=6.32±0.69 |
| | LFSG-NMI | 53.39±2.56 | 48.62±1.77 | 41.51±2.06 | λ=0.2824±0.1426 k=8.64±0.86 |
| USPS | ORACLE-ACC | 75.06±4.81 | 67.73±2.97 | 62.90±3.54 | λ=0.0452±0.1383 k=7.48±1.48 |
| | LFSG-ACC | 70.48±3.67 | 65.98±3.55 **p=0.095** | 60.22±3.66 | λ=0.189±0.1675 k=4.24±2.96 |
| | ORACLE-NMI | 73.64±3.36 | 70.41±1.45 | 64.76±1.68 | λ=0.0012±0.0018 k=5.84±1.14 |
| | LFSG-NMI | 69.90±6.84 **p=0.055** | 64.43±4.75 | 58.72±5.39 | λ=0.1296±0.1377 k=4.92±3.01 **p= 0.069** |
| COIL20 | ORACLE-ACC | 66.07±3.41 | 76.55±2.17 | 60.60±3.49 | λ=0.082±0.1623 k=7.12±1.13 |
| | LFSG-ACC | 64.12±3.23 **p=0.05** | 75.33±2.03 **p=0.044** | 58.50±3.14 | λ=0.1173±0.1289 **p=0.115** k=8±4.67 **p=0.553** |
| | ORACLE-NMI | 66.19±3.58 | 76.74±1.70 | 60.75±3.16 | λ=0.0697±0.1644 k=8.08±0.95 |
| | LFSG-NMI | 64.11±2.57 | 75.12±1.38 | 57.87±2.24 | λ=0.1670±0.1740 k=9.32±7.25 **p= 0.108** |
| COIL100 | ORACLE-ACC | 49.77±1.68 | 75.43±0.87 | 43.99±1.43 | λ=0.0062±0.0198 k=7.27±1.12 |
| | LFSG-ACC | 46.01±3.39 | 72.20±2.68 | 39.67±4.10 | λ=0.0436±0.0852 |
| | | | | | **p=0.555**<br>k=4.09±1.72 |
| | ORACLE-NMI | 49.82±0.98 | 75.98±0.44 | 44.54±0.86 | λ=0.0031±0.0040<br>k=7.68±0.69 |
| | LFSG-NMI | 47.42±2.63 | 73.17±2.17 | 41.02±3.27 | λ=0.0178±0.0704<br>k=5.20±1.50 |

*4.1.4 Sparse SC algorithm*

In the implementation of the sparse SC algorithm, we distinguish between to cases: when the error/noise term has is Gaussian and when it is sparse, see [15]. The first case of the SSC algorithm is derived from equation (5) by setting $f(\mathbf{Z}) = \|\mathbf{Z}\|_1$ and $\tau$=0, requiring the coefficient representation matrix **Z** to be sparse. The second case, robust to outliers, is defined by:

$$\min_{\mathbf{Z}} \|\mathbf{X} - \mathbf{XZ}\|_1 + \lambda \|\mathbf{Z}\|_1 \;\; \text{s.t.} \;\; \text{diag}(\mathbf{Z}) = \mathbf{0}. \qquad (33)$$

Equation (33), represents robust version of the SSC algorithm, used to cluster the extended YaleB dataset. All other datasets were clustered using the non-robust version of the SSC algorithm. The SSC algorithm has one hyperparameter, $\lambda$. In its implementation, as outlined in [15], a scaled regularization constant is introduced $\alpha = \lambda/\mu$. For the non-robust version, $\mu \triangleq \min_i \max_{j \neq i} |\mathbf{x}_i^{\mathrm{T}} \mathbf{x}_j|,\ i, j = 1, \ N$, and for the robust version, $\mu \triangleq \min_i \max_{j \neq i} \|\mathbf{x}_j\|_1,\ i, j = 1, \ N$, giving us $\lambda = \alpha \times \mu$. When defining the search space, we preset $\alpha \in \mathbb{N}_+$. As seen in Table 5, for ORL, EYaleB, MNIST, USPS, and COIL20 the performance of the oracle version on in-sample data is mostly up to 6% better than the LFSG version. For COIL100 dataset, the LFSG version yields up to 2% better performance. On out-of-sample data, the difference in performance is in many cases statistically insignificant.

**Table 5**
The clustering performance of the oracle and LFSG versions of the SSC algorithm. In each version, the hyperparameter α was selected based on either ACC or NMI. Performance metrics were calculated for both in-sample and out-of-sample data. Bold font indicates cases where the difference between the oracle and LFSG versions is statistically insignificant.

| Dataset | ACC [%]<br>in-sample<br>out-of-sample | NMI [%]<br>**i**n-sample | F_score[%]<br>**i**n-sample | Hyperparameter |
|---|---|---|---|---|

| | | | out-of-sample | out-of-sample | |
|---|---|---|---|---|---|
| ORL | ORACLE-ACC | 78.74±1.94<br>70.77±2.54 | 88.90±1.11<br>88.02±1.30 | 68.01±255<br>51.88±4.32 | α=25.56±9.27 |
| | LFSG-ACC | 73.80±1.97<br>71.33±2.70<br>**p_out =0.292** | 86.99±1.00<br>88.00±1.37<br>**p_out =0.801** | 62.55±2.35<br>51.95±4.20<br>**p_out =0.930** | α=28.29±7.54<br>**p=0.260** |
| | ORACLE-NMI | 79.26±2.16<br>71.77±3.09 | 89.18±1.03<br>87.82±1.53 | 68.74±2.78<br>51.95±4.91 | α=30.68±8.42 |
| | LFSG-NMI | 74.40±3.15<br>71.10±3.97<br>**p_out =0.830** | 87.27±1.65<br>87.70±1.76<br>**p_out =0.600** | 62.98±4.23<br>51.14±5.62<br>**p_out =0.628** | α=27.78±8.86<br>**p=0.252** |
| EYaleB | ORACLE-ACC | 76.60±2.14<br>82.15±1.48 | 81.19±1.23<br>87.21±0.82 | 45.53±3.62<br>61.63±3.40 | α=15.44±2.19 |
| | LFSG-ACC | 70.19±3.31<br>78.75±2.51 | 76.47±2.51<br>84.08±1.80 | 28.20±5.84<br>49.54±6.39 | α=33.14±9.00 |
| | ORACLE-NMI | 76.61±0.73<br>82.94±2.19 | 81.36±0.93<br>87.70±1.06 | 48.55±3.86<br>64.40±4.62 | α=15.0±1.33 |
| | LFSG-NMI | 70.34±4.10<br>79.39±3.54 | 76.48±2.61<br>84.60±1.91 | 28.09±5.28<br>51.86±6.77 | α=33.20±6.88 |
| MNIST | ORACLE-ACC | 64.26±4.15<br>60.97±3.57 | 66.67±2.21<br>63.82±2.31 | 55.61±3.43<br>53.32±2.83 | α=5.88±2.54 |
| | LFSG-ACC | 60.29±4.08<br>60.28±3.76<br>**p_out =0.669** | 65.03±3.30<br>63.90±2.80<br>**p_in = 0.068**<br>**p_out = 0.923** | 53.04±3.65<br>53.29±3.28<br>**p_out = 0.786** | α=8.38±1.02 |
| | ORACLE-NMI | 64.38±4.02<br>61.52±3.32 | 65.46±3.51<br>64.81±2.30 | 55.55±3.91<br>53.64±2.54 | α=6.04±2.37 |
| | LFSG-NMI | 62.23±4.04<br>61.37±3.36<br>**p_in = 0.056**<br>**p_out =0.954** | 64.28±2.53<br>64.03±2.46<br>**p_in = 0.214**<br>**p_out =0.162** | 53.72±3.22<br>53.33±2.76<br>**p_in = 0.08**<br>**p_out = 0.846** | α=8.20±1.22 |
| USPS | ORACLE-ACC | 77.46±4.52<br>76.70±5.55 | 75.88±3.47<br>76.06±2.41 | 68.98±4.35<br>69.16±3.40 | α=4.32±1.93 |
| | LFSG-ACC | 72.03±4.80<br>72.60±4.75 | 74.22±3.33<br>74.35±3.17<br>**p_in=0.071** | 65.28±4.74<br>65.90±4.53 | α=7.74±1.42 |
| | ORACLE-NMI | 77.98±4.67<br>73.18±6.20 | 76.712.66<br>75.20±2.85 | 69.32±3.93<br>66.85±5.05 | α=6.48±1.87 |
| | LFSG-NMI | 71.70±4.92<br>71.15±5.50<br>**p_out=0.461** | 74.97±2.93<br>73.81±3.49<br>**p_in = 0.05**<br>**p_out= 0.103** | 65.96±4.23<br>65.14±5.13<br>**p_out=0.207** | α=8.23±1.28 |
| COIL20 | ORACLE-ACC | 80.30±2.79<br>80.33±2.95 | 92.73±1.00<br>92.73±1.37 | 78.33±2.06<br>78.00±2.51 | α=8.84±2.94 |
| | LFSG-ACC | 78.85±3.20<br>79.09±2.94<br>**p_in= 0.103**<br>**p_out=0.168** | 92.40±1.35<br>92.11±1.41<br>**p_in= 0.299**<br>**p_out=0.046** | 78.31±2.19<br>77.41±2.13<br>**p_in= 0.954**<br>**p_out=0.415** | α=11.31±2.69 |
| | ORACLE-NMI | 81.04±2.83<br>81.04±3.12 | 93.13±1.22<br>93.49±0.93 | 78.93±2.28<br>78.84±2.21 | α=8.12±1.42 |
| | LFSG-NMI | 79.12±2.59<br>79.69±2.49<br>**p_out=0.130** | 92.63±0.25<br>92.38±1.01<br>**p_in=0.099** | 78.44±1.71<br>77.68±1.85<br>**p_in = 0.655**<br>**p_out=0.114** | α=11.52±2.13 |
| COIL100 | ORACLE-ACC | 54.13±0.83<br>56.87±0.91 | 81.01±0.51<br>82.69±0.75 | 46.93±0.97<br>49.55±1.89 | α=19.77±7.19 |
| | LFSG-ACC | 54.94±1.59<br>55.82±1.80 | 81.54±0.99<br>81.77±1.03 | 48.87±2.17<br>48.06±2.33 | α=29.37±5.26 |
| | | **p_in = 0.054** **p_out=0.166** | **p_in=0.109** | **p_out=0.126** | |
| | ORACLE-NMI | 54.12±1.09 55.46±1.01 | 81.16±0.29 82.66±0.68 | 47.18±0.63 46.21±3.13 | α=14.78±4.06 |
| | LFSG-NMI | 54.77±1.64 55.87±1.31 **p_in=0.198** **p_out=0.423** | 81.21±1.07 81.57±0.84 **p_in=1** | 48.52±2.08 47.96±1.70 **p_in=0.05** **p_out=0.190** | α=29.32±4.84 |

*4.1.5. Low-rank sparse $S_0 / \ell_0$ SC algorithm*

The S0L0 LRSSC algorithm [16] is derived from equation (5) by setting $f(\mathbf{Z}) = \|\mathbf{Z}\|_{S_0}$ and $g(\mathbf{Z}) = \|\mathbf{Z}\|_0$. In our implementation, as outlined in Algorithm 2 of [16], we further set $\lambda+\tau=1$, meaning $\tau=1-\lambda$. The second hyperparameter, $\alpha>0$, is the penalty constant in the ADMM-based implementation of the S0L0 LRSSC algorithm. As shown in Table 6, for the ORL dataset, the clustering performance of the LFSG version is up to 15% lower than the oracle version, both on in-sample and out-of-sample data. On other datasets, the LFSG version's performance is typically up to 6% lower than that of the oracle version, with similar results observed for out-of-sample data.

**Table 6**
The clustering performance of the oracle and LFSG versions of the S0L0 LRSSC algorithm. In each version, the hyperparameters $\lambda$ and $\alpha$ were selected based on either ACC or NMI. Performance metrics were calculated for both in-sample and out-of-sample data. Bold font indicates cases where the difference between the oracle and LFSG versions is statistically insignificant.

| Dataset | | ACC [%] in-sample out-of-sample | NMI [%] in-sample out-of-sample | F_score[%] in-sample out-of-sample | Hyperparmeters |
|---|---|---|---|---|---|
| ORL | ORACLE-ACC | 71.03±3.29 67.30±3.49 | 85.11±2.40 85.75±2.01 | 57.95±4.31 45.18±6.44 | α=2.68±2.98 λ=0.0341±0.0463 |
| | LFSG-ACC | 57.76±2.98 57.73±3.03 | 76.96±1.50 80.13±1.76 | 38.95±3.83 29.77±4.79 | α=9.06±0.71 λ=0.0029±0.0026 **p=0.318** |
| | ORACLE-NMI | 72.17±2.74 68.27±2.95 | 86.28±1.19 86.33±1.50 | 60.10±3.30 46.78±4.74 | α=2.92±2.97 λ=0.0111±0.027 |
| | LFSG-NMI | 72.17±2.74 68.27±2.95 | 86.28±1.19 86.33±1.50 | 60.10±3.30 46.78±4.74 | α=2.92±2.97 λ=0.0111±0.027 |
| EYaleB | ORACLE-ACC | 89.62±1.66 88.31±2.16 | 90.84±1.15 90.75±1.32 | 78.74±2.69 79.08±2.87 | α=3.56±0.92 λ=0.0027±0.0042 |
| | LFSG-ACC | 83.34±4.77 | 86.77±4.16 | 68.91±9.53 | α=5.85±2.36 |
| | | 84.14±4.07 | 88.00±3.14 | 72.33±7.27 | λ=0.0018±0.0027<br>**p=0.845** |
| | ORACLE-NMI | 89.56±1.66<br>87.92±1.44 | 91.06±0.72<br>90.59±0.81 | 78.97±1.82<br>78.31±1.94 | α=3.4±0.82<br>λ=0.0028±0.0042 |
| | LFSG-NMI | 82.28±5.82<br>83.14±4.92 | 86.22±4.49<br>87.42±3.55 | 67.74±9.42<br>70.60±8.09 | α=6.41±2.74<br>λ=0.0011±0.0019<br>**p=0.241** |
| MNIST | ORACLE-ACC | 64.17±5.01<br>63.44±5.69 | 65.40±2.59<br>63.87±3.48 | 53.95±4.33<br>53.45±4.26 | α=18.84±3.69<br>λ=0.1261±0.2672 |
| | LFSG-ACC | 57.65±4.82<br>57.56±4.95 | 61.39±3.52<br>62.31±2.86<br>**p_out = 0.05** | 50.06±3.58<br>50.58±3.95 | α=17.50±2.54<br>λ=0.0011±0.002 |
| | ORACLE-NMI | 62.47±3.88<br>59.99±5.01 | 65.48±2.59<br>64.88±2.51 | 54.47±3.37<br>53.52±3.54 | α=18.84±2.15<br>λ=0.0317±0.0436 |
| | LFSG-NMI | 59.12±5.52<br>58.96±5.54<br>**p_out =0.600** | 63.89±2.52<br>63.71±3.75<br>**p_in = 0.065**<br>**p_out=0.244** | 52.23±4.55<br>52.02±4.90<br>**p_in = 0.052**<br>**p_out = 0.260** | α=18.81±1.82<br>**p=0.520**<br>λ=0.0006±0.0007 |
| USPS | ORACLE-ACC | 84.03±5.20<br>83.68±5.05 | 78.97±2.97<br>78.65±3.25 | 75.12±4.74<br>74.31±4.79 | α=24.84±4.28<br>λ=0.1013±0.2727 |
| | LFSG-ACC | 80.98±5.45<br>80.92±5.42<br>**p_out= 0.061** | 77.17±2.95<br>77.48±3.05<br>**p_in=0.0625**<br>**p_out=0.1744** | 72.28±4.71<br>72.29±4.37<br>**p_in= 0.05**<br>**p_out= 0.1352** | α=25.48±2.38<br>**p=0.815**<br>λ=0.0004±0.0009 |
| | ORACLE-NMI | 83.85±4.65<br>83.13±6.09 | 79.81±2.46<br>80.06±2.71 | 74.44±3.75<br>75.74±4.30 | α=24.28±3.69<br>λ=0.0235±0.0393 |
| | LFSG- NMI | 78.43±5.94<br>78.83±6.30 | 77.57±3.19<br>78.34±3.19<br>**p_out=0.063** | 71.08±4.75<br>72.00±5.13 | α=23.33±3.84<br>**p=0.437**<br>λ=2.58e-4±2.51e-4 |
| COIL20 | ORACLE-ACC | 78.55±2.71<br>77.18±2.36 | 89.31±1.84<br>88.51±1.91 | 74.82±3.07<br>73.10±2.51 | α=19.57±4.99<br>λ=0.0253±0.0407 |
| | LFSG-ACC | 77.34±2.22<br>77.61±1.7<br>**p_in=0.183**<br>**p_out=0.420** | 87.96±1.99<br>88.19±1.78<br>**p_in=0.063**<br>**p_out=0.663** | 73.48±2.34<br>72.92±1.91<br>**p_in=0.094**<br>**p_out=0.836** | α=24.40±5.54<br>λ=0.003±0.0026<br>**p=0.083** |
| | ORACLE-NMI | 78.33±2.29<br>77.23±3.45 | 89.93±1.71<br>89.52±1.92 | 75.32±2.20<br>73.62±2.42 | α=23.04±5.21<br>λ=0.0273±0.0415 |
| | LFSG-NMI | 77.31±3.27<br>77.22±2.08<br>**p_in=0.133**<br>**p_out=0.862** | 87.74±2.24<br>87.84±2.32<br>**p_out=0.078** | 73.20±2.84<br>72.60±3.16<br>**p_in=0.100**<br>**p_out=0.471** | λ=0.0028±0.0023<br>**p=0.067** |
| COIL100 | ORACLE-ACC | 52.33±0.97<br>53.08±1.32 | 78.19±0.48<br>79.06±1.17 | 45.01±1.74<br>45.34±1.38 | α=25.72±7.44<br>λ=0.18±0.2898 |
| | LFSG-ACC | 47.63±0.64<br>48.67±0.98 | 76.53±0.56<br>77.69±0.57 | 42.78±1.13<br>43.07±1.14 | α=25.90±4.64<br>**p=0.429**<br>λ=0.004±0.003 |
| | ORACLE-NMI | 51.95±1.03<br>52.50±0.68 | 78.50±0.51<br>79.19±0.53 | 45.21±0.89<br>44.80±0.86 | α=28.20±3.28<br>λ=0.1±0.0 |
| | LFSG-NMI | 47.94±0.86<br>48.67±0.83 | 76.73±0.44<br>77.60±0.42 | 42.89±0.78<br>42.74±0.68 | α=29.44±1.59<br>**p=0.5319**<br>λ=0.0015±0.0023 |

### *4.2. Multi-view SC algorithm*

We also evaluated the performance of the proposed approach HPO approach on the multi-view LMVSC algorithm [24] and multi-view MLME algorithm [72]. As there are no out-of-sample extensions for the LMVSC and MLME algorithms, the results presented in Tables 8 and 9 pertain only to in-sample data. We compare clustering performance of MLME and LMVSC algorithms with fast parameter free multi-view subspace clustering algorithm with consensus anchor guidance (FPMVS-CAG), [20], on three multi-view datasets: Handwritten-numerals BBC, and Caltech101-20. They are available at links provided in sections 4.2.1 and 4.2.2. Our motivation is to demonstrate that proposed LFSG version of MLME and LMVSC algorithms outperforms the parameter-free FPMVS-CAG which justifies application of proposed method to the existing SC algorithms.

**Table 7**

Main information on datasets used in the experiments with multi-view SC algorithms. Performance metrics were calculated using in-sample data from 70% randomly selected samples per cluster.

| Dataset | #Sample | #Views | #Cluster |
|---|---|---|---|
| Handwritten numerals | 2000 | 6 | 10 |
| BBC | 685 | 4 | 5 |
| Caltech101-20 | 2386 | 6 | 20 |

*4.2.1. Multi-view LMVSC algorithm*

The LMVSC algorithm, formulated in equation (9), introduces two hyperparameters: the number of anchors $M$ and the regularization constant $\alpha$. The MATLAB code for the original version of this algorithm, provided by the authors, is available at https://github.com/sckangz/LMVSC. Table 8 reports results on 25 runs, using data randomly selected from the 70% of the total number of samples per cluster. As shown in [24] that the optimal hyperparameter values for the Handwritten_numerals dataset are $M$=10 and $\alpha$=$10^{-3}$. In comparison with [24], we slightly modified the search space, $M\in\{10, 15, 25, 50\}$ and $\alpha\in\{10^{-}$

$^{4}$ ,$10^{-3}$, $10^{-2}$, $10^{-1}$, 1, 10}, which improved the oracle's performance compared to the original results in [24]. As shown in Table 8, the oracle version estimated the optimal hyperparameter values $M$=25 and $\alpha$=$10^{-4}$ using either ACC or NMI as the selection metric. The LFSG version estimated hyperparameter values with a slight displacement relative to the oracle values. As can be seen in Table 8, the LFSG version's performance is typically up to 7% lower than that of the oracle version. In comparison with parameter-free FPMVS-CAG algorithms, the LFSG LMVSC algorithm achieves better results on Handwritten numerals and BBC datasets, and worse results on Caltech101-20 dataset. Arguably, the last result could be improved by evaluating the quality of clustering partitions visualization but that would increase computational complexity of LFSG LMVSC method.

**Table 8**

The clustering performance of the oracle and LFGS versions of the LMVSC algorithm. In both versions, the hyperparameters $M$ and $\alpha$ were selected based on either ACC or NMI. Performance metrics were calculated using in-sample data from 70% randomly selected samples per cluster. Bold font is used to highlight cases where the difference between the oracle and LFSG versions is statistically insignificant.

| Dataset | | ACC [%] in-sample | NMI [%] in-sample | F_score[%] in-sample | Hyperparameters |
|---|---|---|---|---|---|
| Handwritten numerals | ORACLE-ACC | 89.57±0.0 | 83.13±0.0 | 79.60±0.0 | α=1e-4±0 M=25±0 |
| | LFSG-ACC | 82.50±0.0 | 78.39±0.0 | 71.92±0.0 | α=4e-4±0 M=18±0 |
| | ORACLE-NMI | 89.10±1.65 | 82.99±0.57 | 79.32±0.86 | α=1e-3±0 M=23.75±4.33 |
| | LFSG-NMI | 79.61±0.97 | 78.21±0.07 | 71.21±0.56 | α=7.42e-4±0 M=16.17±0.58 |
| BBC | ORACLE-ACC | 50.03±0.87 | 23.41±0.46 | 42.88±0.0 | α=10.0±0.0 M=29.0±5.0 |
| | LFSG-ACC | 48.39±0.79 | 21.08±1.46 | 41.02±0.18 | α=5.74±3.67 M=30.2±4.0 |
| | ORACLE-NMI | 42.12±0 | 34.18±0 | 36.99±0 | α=1±0 M=5±0 |
| | LFSG-NMI | 45.23±0 | 23.97±0 | 39.58±8 | α=24.09±0 M=31±0 |
| Caltech101-20 | ORACLE-ACC | 41.57±1.06 | 58.73±1.24 | 34.72±0.57 | α=0.136±0.18 M=52±10 |
| | LFSG-ACC | 39.94±0.07 | 55.91±0.07 | 33.25±0.23 | α=0.0507±0.003 M=86.76±6.2 |
| | ORACLE-NMI | 41.36±0 | 58.98±0 | 34.60±0 | α=0.1±0 M=50±0 |
| | LFSG-NMI | 40.41±0 | 58.50±0 | 33.13±0 | α=0.0794±0<br>M=50±0<br>**p=1** |

*4.2.2 Multi-view MLME algorithm*

The MLME algorithm, formulated in equation (12), introduces two hyperparameters: $\alpha$ and $\beta$. The MATLAB code for the original version of this algorithm, provided by the authors, is available at https://github.com/Ekin102003/MCMLE. We compared the oracle and LFSG versions using the BBC dataset and Handwritten numerals dataset. Table 9 reports results after on 25 runs using data randomly selected from the 70% of the total number of samples per cluster. It was shown in [72] that the optimal values of the hyperparameters for the BBC dataset are $\beta \geq 10$ and $\alpha \in [0.0005, 0.01]$. As shown in Table 9, the oracle version correctly estimated the hyperparameters using either ACC or NMI as the selection metric. The LFSG version also estimated hyperparameter values within the suggested range. The same observation applies to results obtained on the 70% randomly sampled data, although the mean values deviated from those the values obtained on the full dataset.

**Table 9**
The clustering performance of the oracle and LFGS versions of the MLME algorithm. In both versions, the hyperparameters $\alpha$ and $\beta$ were selected based on either ACC or NMI. Performance metrics were calculated using in-sample data from 70% randomly selected samples per cluster. Bold font is used to highlight cases where the difference between the oracle and LFSG versions is statistically insignificant.

| **Dataset** | | **ACC [%]**<br>in-sample | **NMI [%]**<br>**i**n-sample | **F_score[%]**<br>**i**n-sample | **Hyperparmeters** |
|---|---|---|---|---|---|
| BBC | ORACLE-ACC | 84.98±6.97 | 71.82±7.51 | 78.15±8.01 | α=0.003±0.004 |
| | | | | | β=205.4±406.0 |
| | LFSG-ACC | 79.23±10.65 | 68.87±8.51 **p=0.178** | 73.88±10.21 **p=0.112** | α=0.0019±0.0025 **p=0.231** β=55.99±151.17 |
| | ORACLE-NMI | 83.33±5.82 | 73.29±4.36 | 78.51±5.15 | α=0.0023±0.0035 β=129.56±329.18 |
| | LFSG-NMI | 79.55±5.52 | 70.11±4.87 | 75.32±5.08 | α=0.0012±0.0017 β=21.53±31.85 |
| Handwritten numerals | ORACLE-ACC | 88.96±4.34 | 88.24±2.82 | 85.36±4.36 | α=0.0017±0.0028 β=80±236.78 |
| | LFSG-ACC | 80.19±12.57 | 83.03±11.59 | 77.71±12.95 | α=0.0081±0.0225 β=103.74±119.78 |
| | ORACLE-NMI | 86.28±8.60 | 88.04±3.26 | 84.32±6.46 | α=0.0008±0.0004 β=159.64±299.35 |
| | LFSG-NMI | 83.06±9.37 **p=0.269** | 86.75±3.42 **p=0.125** | 81.80±6.61 **p=0.187** | α=0.0006±0.0002 β=198.49±211.14 |

*4.2.3 Fast parameter-free multi-view subspace clustering with consensus anchor guidance*

The fast parameter-free multi-view subspace clustering with consensus anchor guidance method (FPMVS-CAG) is proposed in [20], with the MATAB code available at: https://github.com/wangsiwei2010/FPMVS-CAG. In this algorithm anchor selection and subspace graph construction are conducted into a unified optimization framework. Importantly, the algorithm is proven to have linear time complexity with respect to the number of samples. Even more important, the algorithm can learn the anchor-graph structure without hyperparameters.

**Table 10**

The clustering performance of the FPMVS-CAG algorithm. Performance metrics were calculated using in-sample data from 70% randomly selected samples per cluster.

| Dataset | ACC [%] in-sample | NMI [%] in-sample | F_score[%] in-sample |
|---|---|---|---|
| Handwritten_numerals | 82.25±0 | 79.30±0 | 75.60±0 |
| BBC | 32.26±0 | 2.97±0 | 27.59±0 |
| Caltech101-20 | 65.47±0 | 63.26±0 | 69.05±0 |

### *4.3 Discussion of strengths and weaknesses*

We propose a label-free, self-guided HPO method for SC algorithms. This approach relies on clustering quality metrics such as ACC or NMI, which are computed from pseudo-labels generated by the SC algorithm itself. Similar to the grid-search method in traditional HPO, we define a search space for selected SC algorithm, where the optimal hyperparameters are likely to be located. To evaluate the hyperparameters, we compute ACC or NMI between pseudo-labels generated for neighboring hyperparameter values. Leveraging the smoothness assumption, we iteratively subdivide hyperparameter intervals into smaller sections, which are then further split into halves or thirds. The SC algorithm generates pseudo-labels for these interpolated values. This process is repeated until a predefined relative error criterion is met. To the best of our knowledge, SC algorithms also face challenges with interpretability. To address this, we introduce a method for interpreting clustering results by visualizing subspace bases derived from the clustering partitions. The main strengths of proposed label-free, self-guided SC method include: (i) enabling the re-use of existing SC algorithms in domains without annotated data, and (ii) providing explanations for decisions made by selected SC algorithm. If experts find the visualization quality inadequate, the initial hyperparameter search space can be refined, and the optimization process restarted. However, the proposed method has certain limitations. It critically depends on: (i) the validity of the smoothness assumption for ACC and NMI clustering performance metrics, and (ii) the quality of the initial assessment of the hyperparameter search space. To reduce computational complexity, a less dense search space is preferred. However, if ACC or NMI metrics exhibit oscillatory behavior are on a smaller scale, the hyperparameter search space must be redefined. While this can be achieved by evaluating the quality of clustering partitions visualization, doing so increases computational complexity.

## 5.0 Conclusion

Numerous subspace clustering algorithms have been developed, many of which achieve excellent clustering performance when external labels are available for hyperparameters tuning. However, in many application domains number of unlabeled datasets is growing, and labeling them remains a costly and time consuming task. To address this challenge, one research direction focuses on designing SC algorithms that are inherently free of hyperparameters, thus eliminating the need for HPO. For hyperparameters-dependent SC algorithms, a label-independent approach to HPO involves the use of internal clustering quality metrics (if available). Ideally, these metrics should correlated strongly with external, label-depends metrics. A third approach is the development of entirely hyperparameter-free subspace clustering algorithms. However, their number such algorithms is limited, and their performance often falls short compared to that of hyperparameter-tuned SC algorithms. Our proposed method, the LFSG SC approach, addresses label-independent HPO by utilizing clustering quality metrics such as ACC or NMI, computed from pseudo-labels generated by the SC algorithm over a predefined grid of hyperparameters. By assuming that ACC (or NMI) behaves as a smooth function of hyperparameters, we iteratively refine the hyperparameter intervals by selecting promising subintervals until convergence. A potential weakness of proposed methodology lies in the initial assessment of the hyperparameter search space. To mitigate this issue, we suggest visualizing the clustering partitions generated by the LFSG SC algorithm. This visualization enables domain experts to evaluate clustering quality and, if necessary, guide the refinement of the hyperparameters search space. We anticipate that the LFSG SC method will facilitate repurposing existing SC algorithms in new domains with unlabeled data, overcoming the challenge of label dependency.

**Declaration of Competing Interests**

We declare that this manuscript is original, has not been published before and is not currently being considered for publication elsewhere. We declare that we do not have any known competing financial interests of personal relationships that could have appeared to influence the work reported in this paper.

**Data Availability**

Data used in reported experiments are available either at links cited in the paper or at: https://github.com/ikopriva/LFSGSC.

**Acknowledgments**

This work was supported by grant IP-2022-10-6403 funded by the Croatian Science Foundation.

**References**

[1] A. E. Ezugwu, A. M. Ikotun, O. A. Oyelade, et al., A comprehensive survey of clustering algorithms: State-of-the-art machine learning applications, taxonomy, challenges, and future research prospects, Eng. Appl. Art. Intell. 110 (2022) 104473.

[2] C.-N. Jiao, J. Shang, F. Li, et al., Diagnosis-Guided Deep Subspace Clustering Association Study for Pathogenetic Markers Identification of Alzheimer's Disease Based on Comparative Atlases, IEEE J. Biomed. Health Inform. 28 (5) (2024) 3029-3041.

[3] N. Kumar, P. Uppala, K. Duddu, et al., Hyperspectral Tissue Image Segmentation Using Semi-Supervised NMF and Hierarchical Clustering, IEEE Trans. Med. Imag. 38 (5) (2019) 1304-1313.

[4]. A. F. Møller, J. G. S. Madsen, JOINTLY: interpretable joint clustering of single-cell transcriptomes, Nature Comm. 14 (2023) 8473.

[5] J. Tang, L. Jin, Z. Li, S. Gao, RGB-D object recognition via incorporating latent data structure and prior knowledge, IEEE Trans. Multimed. 17 (11) (2015) 1899–1908.

[6] S. V. Ault, R. J. Perez, C. A. Kimble, J. Wang, On speech recognition algorithms, Int. J. Mach. Learn. Comput. 8 (6) (2018) 518-523.

[7] J. Shen, X. Hao, Z. Liang, et al., Real-time superpixel segmentation by dbscan clustering algorithm, IEEE Trans. Image Proc. 25 (12) (2016) 5933-5942.

[8] W. Wu, M. Peng, A data mining approach combining *k*-means clustering with bagging neural network for short-term wind power forecasting, IEEE Int. Things J. 4 (4) (2017) 979-986.

[9] J. A. Hartigan, M. A.Wong, Algorithm as 136: A k-means clustering algorithm, J. Roy. Stat. Soc. 28 (1) (1979) 100–108.

[10] A. Y. Ng, M. I. Jordan, Y. Weiss, On spectral clustering: Analysis and an algorithm, in: Proc. Adv. Neural Inf. Process. Syst., 2001, pp. 849–856.

[11] U. von Luxburg, A tutorial on spectral clustering, Stat. Comput. 17 (4) (2007) 395–416.

[12] L. Ding, C. Li, D. Jin, S. Ding, Survey of spectral clustering based on graph theory, Pattern Recognit. (151) (2024) 110366.

[13] R. Vidal, Subspace clustering, IEEE Sig. Process. Mag. 28 (2) (2011) 52-68, 2011.

[14] G. Liu, Z. Lin, S. Yan, et al., Robust recovery of subspace structures by low-rank representation, IEEE Trans. Pattern Anal. Mach. Intell. 35 (1) (2013) 171–184.

[15] E. Elhamifar, R. Vidal, Sparse Subspace Clustering: Algorithm, Theory, and Applications, IEEE Trans. Pattern Anal. Mach. Intell. 35 (1) (2013) 2765-2781.

[16] M. Brbić, I. Kopriva, $\ell_0$ Motivated Low-Rank Sparse Subspace Clustering, IEEE Trans. Cyber. 50 (4) (2020) 1711-1725.

[17] C.-Y. Lu, H. Min, Z.-Q. Zhao, et al., Robust and efficient subspace segmentation via least squares regression, in: Computer Vision-ECCV 2012, 2012, pp. 347–360.

[18] Y. Du, G. F. Lu, G. Ji, Robust Least Squares Regression for Subspace Clustering: A Multi-View Clustering Perspective, IEEE Trans. Image Process. 33 (2024) 216-227.

[19] M. Brbić, I. Kopriva, Multi-view Low-rank Sparse Subspace Clustering, Pattern Recognit. 73 (2018) 247-268.

[20] S. Wang, X. Liu, X. Zhu, et al., Fast Parameter-free Multi-view Subspace Clustering with Consensus Anchor Guidance, IEEE Trans. Image Process. 31 (2022) 556-568.

[21] X. Li, H. Zhang, R. Wang, F. Nie, Multiview Clustering: A Scalable and Parameter-Free Bipartite Graph Fusion Method, IEEE Trans. Pattern Anal. Mach. Intell. 44 (1) (2022) 330-344.

[22] Z. Chen, X. J.Wu, T. Xu, J. Kittler, Fast Self-Guided Multi-View Subspace Clustering, IEEE Trans. Image Process. 32 (2023) 6514-6525.

[23] C. Tang, M. Wang, K. Sun, One-Step Multiview Clustering via Adaptive Graph Learning and Spectral Rotation, IEEE Trans. Neural Netw. Learn. Syst. DOI: 10.1109/TNNLS.2024.3381223

[24] Z. Kang, W. Zhu, Z. Zhao, et al., Large-Scale Multi-View Subspace Clustering in Linear Time, in: Proc. of The Thirty Four AAAI Conference on Artificial Intelligence (AAAI-2020), 2020, pp. 4412-4419.

[25] V. M. Patel, R. Vidal, Kernel sparse subspace clustering, in: IEEE International Conference on Image Processing (ICIP), 2014, pp. 2849–2853.

[26] S. Xiao, M. Tan, D. Xu, Z.Y. Dong, Robust kernel low-rank representation, IEEE Trans. Neural Netw. Learn. Syst. 27 (11) (2015) 2268–2281.

[27] Y. Xie, J. Liu, Y. Qu, et al., Robust Kernelized Multiview Self-Representation for Subspace Clustering, IEEE Trans. Neural Netw. Learn. Syst. 32 (3) (2021) 868-881.

[28] Z. Ma, Z. Kang, G. Luo, et al., Towards Clustering-friendly Representations: Subspace Clustering via Graph Filtering, in: Proceedings of the Media Interpretation & Mobile Multimedia (MM'20), 2020, pp. 3081-3089.

[29] Z. Lin, Z. Kang, Graph Filter-based Multi-view Attributed Graph Clustering, in: Proceedings of the Thirtieth International Joint Conference on Artificial Intelligence (IJCAI-21), 2021, pp. 2723-2729.

[30] Z. Lin, Z. Kang, L. Zhang, L. Tian, Multi-view Attributed Graph Clustering, IEEE Trans. Knowl. Data Eng. 35 (2) (2023) 1872-1880.

[31] P. Ji, T. Zhang, H. Li, et al, Deep subspace clustering networks,” in: Proc. Adv. Neural Inf. Process. Syst., 2017, pp. 24–33.

[32] Z. Peng, Y. Jia, H. Liu, et al., Maximum Entropy Subspace Clustering, IEEE Trans. Circ. Syst. Video Tech. 32 (4) (2022) 2199-2210.

[33] Z. Peng, H. Liu, Y. Jia, J. Hou, Adaptive Attribute and Structure Subspace Clustering Network, IEEE Trans. Image Proc. 31 (2022) 3430-3439.

[34] J. Lv, Z. Kang, X. Lu, Z. Xu, Pseudo-supervised deep subspace clustering, IEEE Trans. Image Proc. 30 (2021) 5252-5263.

[35] Y. Xu, S. Chen, J. Li, et al., Autoencoder-Based Latent Block-Diagonal Representation for Subspace Clustering, IEEE Trans. Cyber. 52 (6) (2022) 5408-5418.

[36] K. Li, H. Liu, Y. Zhang, et al., Self-Guided Deep Multi-view Subspace Clustering via Consensus Affinity Regularization, IEEE Trans. Cyber. 52 (12) (2022) 12734-12744.

[37] P. A. Traganitis, G. B. Giannakis, Sketched Subspace Clustering, IEEE Trans. Sig. Proc. 66 (7) (2018) 1663-1675.

[38] J. Xu , Y. Ren , H. Tang, et al., Self-Supervised Discriminative Feature Learning for Deep Multi-View Clustering, IEEE Trans. Knowl. Data Eng. 35 (7) (2023) 7470-7482.

[39] J. Gui, T. Chen, Q. Cao, et al., A Survey of Self-Supervised Learning from Multiple Perspectives: Algorithms, Theory, Applications and Future Trends, IEEE Trans. Pattern Anal. Mach. Intell. DOI: 10.1109/TPAMI.2024.3415112

[40] C. Reed, S. Metzger, A. Srinivas, et al., Evaluating Self-Supervised Pretraining Without Using Labels, (2020) arXiv:2009.07724v1.

[41] N. Huang, L. Xiao, Q. Liu, J. Chanussot, $S^2$DMSC: A Self-supervised Deep Multi-level Subspace Clustering Approach for Large Hyperspectral Images, IEEE Trans. Geosc. Remote Sens. 61 (2023) 5512817.

[42] S. Gidaris, P. Singh, N. Komodakis, Unsupervised representation learning by predictive image rotations, in: International Conference on Learning Representations, 2018, pp. 1-14.

[43] J. Lipor, L. Balzano, Clustering quality metrics for subspace clustering, Pattern Recognit. 104 (2020) 107328.

[44] O. Arbelaitz, I. Gurrutxaga, J. Muguerza, et al., An extensive comparative study of cluster validity indices, Pattern Recognit. 46 (2013) 243-256.

[45] T. Chakraborty, A. Dalmia, A. Mukherjee, N. Ganguly, Metrics for community analysis: a survey, ACM Comput. Surv. (CSUR). 50 (4) (2017) 54.

[46] B. Desgraupes, Clustering indices, Univ. Paris Quest Lab Modal X. (2103) 1-34.

[47] T. A. Bailey Jr., R. Dubes, Cluster validity profiles, Pattern Recognit. 15 (2) (1982) 61-83.

[48] P. S. Bradley, O. L. Mangasarian, "*k*-plane clustering," J. Global Optim. 16 (2000) 23-32.

[49] P. Tseng, "nearest q-flta to m points," J. Optim. Theory Appl. 105 (1) (2000) 249-252.

[50] P.K. Agarwal, N. H. Mustafa, "K-means projective clustering," in: Proceedings of the twenty-third ACM SIGMOD-SIGACT-SIGART Symposium on Principles of Database systems, 2004, pp. 155-165.

[51] T. Caliński, J. Harabasz, A dendrite method for cluster analysis, Commun. Stat. Theory Meth. 3 (1) (1974) 1-27.

[52] L. Deng, M. Xiao, A New Automatic Hyperparameter Recommendation Approach Under Low-Rank Tensor Completion Framework, IEEE Trans. Pattern Anal. Machine Intell. 45 (4) (2023) 4038-4050.

[53] J. Bergstra, Y. Bengio, Random search for hyper-parameter optimization, J. Mach. Learn. Res. 13 (1) (2012) 281–305.

[54] J. Snoek, H. Larochelle, R. P. Adams, Practical Bayesian optimization of machine learning algorithms, in: Proc. Adv. Neural Inf.. Process. Syst., 2012, pp. 2951–2959.

[55] L. Li, K. Jamieson, G. DeSalvo, et al., Hyperband: A novel bandit-based approach to hyperparameter optimization, J. Mach. Learn. Res. 18 (1) (2017) 6765–6816.

[56] H. Mühlenbein, G. Paass, From recombination of genes to the estimation of distributions I. Binary parameters, in: Proc. Int. Conf. Parallel Problem Solving Nat., 1996, pp. 178–187.

[57] X. Guo, J. Yang, C. Wu, et al., A novel LS-SVMs hyper-parameter selection based on particle swarm optimization, Neurocomput. 71 (16-18) (2008) 3211–3215.

[58] J. Vanschoren, Meta-learning, in: Proc. Automa. Mach. Learn., 2019, pp. 35–61.

[59] S. Gandy, B. Recht, I. Yamada, Tensor completion and low-rank tensor recovery via convex optimization, Inverse Probl. 27 (2) (2011) 025010.

[60] J. Liu, P. Musialski, P. Wonka, J. Ye, Tensor completion for estimating missing values in visual data, IEEE Trans. Pattern Anal. Mach. Intell. 35 (1) (2012) 208–220.

[61] P. Zhou, C. Lu, Z. Lin, C. Zhang, Tensor factorization for low-rank tensor completion, IEEE Trans. Image Process. 27 (3) (2017) 1152–1163.

[62] K. Ghasedi Dizaji, A. Herandi, C. Deng,  et al., Deep clustering via joint convolutional autoencoder embedding and relative entropy minimization, in: International Conference on Computer Vision (ICCV), 2017, pp. 5736–5745.

[63] J. Xie, R. Girshick, A. Farhadi, Unsupervised deep embedding for clustering analysis, in: Int. Conf. Mach. Learn. (ICML), 2016, pp. 478–487.

[64] Z. Hao, Z. Lu, G. Li, et al., Ensemble Clustering with Attentional Representation, IEEE Trans. Knowl. Data Eng. 36 (2) (2024) 581-592.

[65] C. Rudin, Stop explaining black box machine learning models for high stakes decisions and use interpretable models instead, Nature Mach. Intell. 1 (2019) 206-215.

[66]. E. Toja, C. Guan, A Survey on Explainable Artificial Intelligence (XAI): Toward Medical XAI,  IEEE Trans. Neural. Netw. Learn. Syst. 32 (11) (2021), 4793-4813.

[67] S. Boyd, N. Parikh, E. Chu, et al., Distributed optimization and statistical learning via the alternating direction method of multipliers, Found. Trends Mach. Learn. 3 (1) (2011) 1–122.

[68] Z. Ma, Z. Kang, G. Luo, et al, Towards Clustering-friendly Representations: Subspace Clustering via Graph Filtering, in: Proceedings of the Media Interpretation & Mobile Multimedia (MM'20), 2020, pp. 3081-3089.

[69] W. Ju, D Xiang, B. Zhang, et al., Random Walk and Graph Cut for Co-Segmentation of Lung Tumor on PET-CT Images with 3D Derivative Features, IEEE Trans. Image Proc. 29 (12) (2015) 5854-5867.

[70] M. Amini, N. Usunier, C. Goutte, Learning from multiple partially observed views - an application to multilingual text categorization, in: Proc. Adv. Neural Inf. Process. Syst., 2009, pp. 28–36.

[71] W. Hao, S. Pang, B. Yang, J. Xue, Tensor-based multi-view clustering with consistency exploration and diversity regularization, Knowl. Based Syst. 252 (2022), 109342.

[72] G. Zhong, C. M. Pun, Improved Normalized Cut for Multi-View Clustering, IEEE Trans. Pattern Anal. Mach. Intell. 44 (12) (2022) 10244-10251.

[73] Z. Wang, Z. Li, R. Wang, et al., Large graph clustering with simultaneous spectral embedding and discretization, IEEE Trans. Pattern Anal. Mach. Intell. 43 (12) (2020) 4426-4440.

[74] N. Kwak, Nonlinear Projection Trick in Kernel Methods: An Alternative to the Kernel Trick, IEEE Trans. Neural Netw. Learn. Syst. 24 (12) (2013) 2113-2119.

[75] K.-C. Lee, J. Ho, D. Kriegman, Acquiring linear subspaces for face recognition under variable lighting, IEEE Trans. Pattern Anal. Mach. Intell. 27 (5) (2005) 684–698.

[76] T. Hastie, P. Y. Simard, Metrics and models for handwritten character recognition, Statist. Sci. 13 (1) (1998) 54–65.

[77] Y. LeCun, L. Bottou, Y. Bengio, P. Haffner, Gradient-based learning applied to document recognition, Proc. of The IEEE 86 (11) (1998) 2278–2324.

[78] J. J. Hull, A database for handwritten text recognition research, IEEE Trans. Pattern Anal. Mach. Intell. 16 (5) (1994) 550–554.

[79] A. S. Georghiades, P. N. Belhumeur, D. J. Kriegman, From few to many: Illumination cone models for face recognition under variable lighting and pose, IEEE Trans. Pattern Anal. Mach. Intell. 23 (6) (2001) 643–660.

[80] F. S. Samaria, A. C. Harter, Parameterisation of a stochastic model for human face identification, in: Proc. 1994 IEEE Workshop on Applic. Comp. Vis. (WACV IEEE), 1994, pp. 138–142.

[81] S. A. Nene, S. K. Nayar, H. Murase, Columbia object image library (coil-100), Tech. Report, CUCS-006-96, Dept. of Computer Science, Columbia Univ, 1996, https://www1.cs.columbia.edu/CAVE/software/softlib/coil-100.php.

[82] D. Greene, P. Cunningham, Practical solutions to the problem of diagonal dominance in kernel document clustering, in: *Proc. Int. Conf. Mach. Learn.*, 2006, pp. 377–384.